\documentclass[10pt,journal,compsoc]{IEEEtran}
\usepackage{amsmath,amssymb,amsfonts}
\usepackage{graphicx}
\usepackage{multirow}
\usepackage{booktabs}

\usepackage{url}
\usepackage[switch]{lineno}  

\newif\iflinenumbers

\iflinenumbers
  \linenumbers
  \modulolinenumbers[1] 
\fi
\usepackage{color}
\usepackage{hyperref}
\hypersetup{colorlinks=true, linkcolor=blue, citecolor=blue, urlcolor=blue}
\usepackage{caption}
\usepackage[nocompress]{cite}
\usepackage[numbers]{natbib}
\makeatletter
\renewcommand{\NAT@separator}{,}

\makeatother
\usepackage[caption=false,font=normalsize,labelfont=sf,textfont=sf]{subfig}
\usepackage{amsthm} 
\newtheorem{theorem}{Theorem} 

\usepackage{mathtools}
\usepackage{mathrsfs} 
\usepackage{multirow} 
\usepackage{textcomp} 
\usepackage{booktabs} 
\usepackage{algorithm} 
\usepackage{algorithmicx} 
\usepackage{algpseudocode} 
\usepackage{listings} 
\usepackage{CJKutf8} 
\usepackage{adjustbox} 
\usepackage{pifont} 
\PassOptionsToPackage{section}{float}
\usepackage{float} 
\usepackage[table]{xcolor} 
\definecolor{hlcore}{RGB}{255,255,153} 
\definecolor{hlcr}{RGB}{200,230,201}     
\definecolor{hldelta}{RGB}{187,222,251}  
\definecolor{hlall}{RGB}{255,205,210}    
\usepackage{placeins} 
\usepackage{hhline}
\usepackage{tabularx}
\makeatletter
\newcommand{\setref}[2]{%
  \global\@namedef{r@#1}{{#2}{}}%
}
\makeatother
\setref{app:proofs}{S1}
\setref{app:proofs T1}{S1.2}
\setref{lem:com}{1}
\setref{eq:cs_appendix}{S3}
\setref{lem:pb_source}{2}
\setref{eq:source_pb}{S6}
\setref{eq:com_apply}{S7}
\setref{app:pacbayes_local_prior}{S1.3}
\setref{eq:dv_var}{S9}
\setref{eq:dv_ineq}{S10}
\setref{eq:Et_def}{S11}
\setref{eq:Pt_def}{S12}
\setref{eq:kl_decomp}{S13}
\setref{app:robustness}{S2}
\setref{app:add res for geo deg}{S2.1}
\setref{fig:stress_res_add}{S1}
\setref{fig:stress_icdm}{S2}
\setref{app:ece}{S2.2}
\setref{app:icdm}{S2.3}
\setref{fig:visda}{S3}
\setref{app:imbalance_setup}{S2.4}
\setref{fig:stat_diff_class_dis}{S4}
\setref{tab:imb_patterns_used}{S1}
\setref{app:label_mismatch}{S2.5}
\setref{fig:class_wise_acc}{S5}
\setref{app:MACW}{S3}
\setref{fig:class_imbalanced_summary}{S6}
\setref{app: ablation FT}{S4}
\setref{tab:margin_weight_formulas}{S2}
\setref{tab:weighting_ablation_results}{S3}
\setref{tab:controlled_complexity_pa}{S4}
\setref{app: polar visual}{S5}
\setref{tab:hyperparam_settings}{S5}
\setref{fig:difference_w_factor}{S7}
\setref{app: add Feat Anal}{S6}
\setref{fig:label_mismatch}{S8}
\setref{app: ESAT_RGB}{S7}
\setref{fig:fint_hyp}{S9}
\setref{selossmap}{S10}
\setref{normlizedresnet}{S11}
\setref{ESAT_RGB}{S12}
\setref{app: Eq Det}{S8}
\setref{eq_label_pre}{S13}
\setref{app: Prot Struct}{S9}
\setref{app:hypersetting-benchmark}{S10}
\setref{app: real-world hypersetting}{S11}
\setref{tab:hyperparameters-final}{S6}
\setref{tab:hyperparameters-multimodal}{S7}
\setref{tab:hyperparameters_app}{S8}
\setref{tab:Append_grid_visda}{S9}
\setref{tab:Append_grid_clipart2product}{S10}
\setref{fig:alpha on decision booundaries}{S14}
\setref{fig:beta on decision booundaries}{S15}

\title{Feature-Space Planes Searcher: A Universal Domain Adaptation Framework for Interpretability and Computational Efficiency}

\author{Zhitong Cheng+, Yiran Jiang+, Yulong Ge, Yufeng Li, Zhongheng Qin, Rongzhi Lin, Jianwei Ma 
\thanks{This work was supported in part by the Open Fund Project of Sinopec Geophysical Research Institute Co., Ltd. under Grant 36750000-24-FW0399-0009, in part by the China Postdoctoral Science Foundation under Grant 2023M740902, in part by the National Natural Science Foundation of China under Grant 42230806 and Grant U23B6010, and in part by the China National Petroleum Corporation--Peking University Strategic Cooperation Project of Fundamental Research.}
\thanks{Z. Cheng, Y. Jiang, Y. Ge, Y. Li, Z. Qin, and R. Lin are with the School of Mathematics and Institute for Artificial Intelligence, Harbin Institute of Technology, China.}
\thanks{J. Ma is with the School of Mathematics and Institute for Artificial Intelligence, Harbin Institute of Technology, China and the School of Earth and Space Sciences, Institute for Artificial Intelligence, Peking University, China.}
\thanks{Corresponding author: jwm@pku.edu.cn}
\thanks{+: These authors contributed equally.}
}

\begin{document}
\iflinenumbers
  \linenumbers
\fi
\IEEEtitleabstractindextext{%

\begin{abstract}
Domain shift, characterized by degraded model performance during the transfer from labeled source domains to unlabeled target domains, poses a persistent challenge for deploying deep learning systems. Current unsupervised domain adaptation (UDA) methods predominantly rely on fine-tuning feature extractors—an approach limited by high computational cost, reduced interpretability, and poor scalability to modern architectures. Our analysis reveals that models pre-trained on large-scale data exhibit domain-invariant geometric patterns in their feature space, characterized by intra-class clustering and inter-class separation, thereby preserving transferable discriminative structures. These findings suggest that cross-domain performance degradation is often associated with decision-boundary misalignment, and that correcting such misalignment can serve as an effective alternative to feature adaptation, particularly when pretrained representations are sufficiently strong. Unlike fine-tuning entire pre-trained models, which risks introducing unpredictable feature distortions, we propose the Feature-space Planes Searcher (FPS): a novel domain adaptation framework that optimizes decision boundaries by leveraging these geometric patterns while keeping the feature encoder frozen. This streamlined approach enables interpretable analysis of adaptation while substantially reducing memory and computational costs through offline feature extraction, permitting full-dataset optimization in a single training cycle. 
Moreover, we introduce an Intra-Class Distance Metric (ICDM) that enables fully unsupervised hyperparameter selection without requiring target-domain labels.
Evaluations on public benchmarks show that FPS achieves competitive performance across standard benchmarks, with notable gains in several settings and tasks. FPS scales efficiently with large multimodal  models and shows versatility across diverse domains including protein structure prediction, remote sensing classification, and earthquake detection. 
We anticipate FPS will provide a simple, effective, and generalizable framework for domain adaptation tasks.
\end{abstract}

}

\maketitle
\IEEEdisplaynontitleabstractindextext
\IEEEpeerreviewmaketitle

\IEEEraisesectionheading{\section{INTRODUCTION}}
\IEEEPARstart{D}espite the transformative impact of deep learning across diverse domains, a key limitation persists: the dependence on large-scale labeled datasets to adequately represent sample spaces ~\cite{kirillovSegmentAnything2023a,openaiGPT4TechnicalReport2023}. Acquiring such data remains costly and labor-intensive~\cite{liaoSuperresolutionStrategyMass2023}. This challenge is further amplified in real-world scenarios, where domain shift occurs because of significant discrepancies between training and application data distributions. The objective of unsupervised domain adaptation (UDA) is to enhance the generalization capability of existing models on new, completely unlabeled data, serving as a crucial approach to improving the practical applicability of corresponding methods. 

Researchers have proposed diverse UDA techniques, including strategies based on domain alignment, consistency regularization, and self-supervised learning~\cite{ganin2016domain, Zhang2018CollaborativeAA,Chen2020AdversarialLearnedLF, song2023ecotta, Liang2020DoWR, prabhu2021sentry, Zou2019ConfidenceRS, Lai2023PADCLIPPW, Na2021ContrastiveVS, westfechtel2024gradual}. With the rise of vision transformers, there have been some tailored methodologies, which leverage Transformer’s inductive bias to innovatively design feature-mixing and patch-mixing mechanisms, as well as exploit the inherent transferability of its structure \cite{xu2021cdtrans, yu2024feature, zhu2023patch, yang2023tvt}. Concurrently, recent advancements in pre-trained multimodal large models \cite{singha2024adclipadaptingdomainsprompt, Westfechtel2023CombiningIK, Bai2023PromptbasedDA, Chen2024EmpoweringSD} have driven significant performance and efficiency gains in UDA. Through lightweight fine-tuning or innovative prompting approaches, researchers optimize pre-trained models for domain adaptation while minimizing feature degradation \cite{Zhou2024UnsupervisedDA, Gao2021CLIPAdapterBV, Bai2023PromptbasedDA, laiPADCLIPPseudolabelingAdaptive2023}.  These advances have significantly improved cross-domain generalization performance.  However, most existing approaches still rely on adapting the feature extractor to reduce discrepancies between source and target domains.

While feature fine-tuning has proven effective, it often requires updating large numbers of parameters.
Since the adaptation dataset is typically much smaller than the pretraining dataset, this discrepancy may lead to under-constrained parameter updates and increased computational cost \cite{Zheng2025}.
These challenges become even more pronounced for modern multimodal foundation models.
Moreover, updating large neural networks during adaptation makes the transfer process harder to interpret and control. Additionally, except for a few studies \cite{Na2021ContrastiveVS} considering the specific distribution, most approaches focus solely on the total discrepancy between source and target domain distributions, preventing the prior distribution from being fully leveraged as a crucial constraint. Moreover, analyzing the overall distribution of target domain features is computationally prohibitive, making such research even rarer.

Motivated by this question, we examine the geometric structure of pretrained feature spaces.
We observe consistent geometric patterns characterized by intra-class clustering and inter-class separation in the frozen feature space of pretrained models, implying the existence of a shared cross-domain decision boundary. This observation suggests that correcting decision-boundary misalignment within the frozen feature space may improve cross-domain performance without modifying the feature representations. Accordingly, unsupervised domain adaptation can be formulated as refining the decision boundary while keeping the feature extractor frozen. The problem thus reduces to identifying constraints for fine-tuning decision planes, circumventing uncertain feature variations introduced by adjusting the entire pre-trained model. Leveraging this insight and domain-invariant geometric patterns, we propose a novel framework for domain adaptation: the Feature-Space Planes Searcher (FPS). FPS optimizes decision planes by maximizing sample discriminability while minimizing class imbalance in the target domain. Within the fixed feature space, this task simplifies to extending decision boundaries from source-domain to target-domain regions, guided by prior knowledge. This framework enables rigorous examination of prior assumptions and their impact on the transfer processes. By pre-extracting features, memory and computational costs are reduced, facilitating single-step optimization across the full dataset distribution.

Comprehensive evaluations on unsupervised domain adaptation benchmarks (Office-31 \cite{2010Adapting}, Office-Home \cite{2017Deep}, and VisDA-2017 ~\cite{Peng2017VisDATV}) demonstrate that FPS achieves competitive performance relative to advanced methods, with notable improvements in certain scenarios. 
In addition, FPS exhibits enhanced class balance capabilities, effectively addressing the issue of disproportionately low classification accuracies for specific categories. Our PAC-style theoretical~\cite{pmlr-v48-germain16} analysis motivates the hypothesis that, under the same optimization constraints, restricting adaptation to the decision boundary may generalize more favorably than fine-tuning larger portions of the network; the controlled experiments in Section \ref{fint_abl} and Table \ref{tab:extent_controlled_comparison} provide supporting evidence for this trend.

FPS also benefits from stronger pretrained encoders, and its frozen-feature design makes it naturally compatible with large multimodal pretrained models.
Leveraging the stability of the feature space, we introduce geometric metrics such as average intra-class distance for hyperparameter optimization in the absence of target domain labels.
Practical applicability is demonstrated through successful deployment in real-world tasks including protein structure prediction, remote sensing image classification, and earthquake event detection, highlighting FPS's substantial potential for cross-domain applications.
\subsection*{Contributions}
Our main contributions are:
\begin{enumerate}
\item \textbf{Revisiting domain adaptation in pretrained feature spaces.}
We investigate the geometric structure of pretrained feature spaces and observe that they often exhibit consistent intra-class clustering and inter-class separation across domains.
This observation suggests that cross-domain performance can often be improved by refining decision boundaries within a frozen feature space, without necessarily modifying the feature representations.

  \item \textbf{Decision-plane adaptation via structure-informed priors in a frozen feature space.}
  We propose \emph{Feature-Space Planes Searcher} (FPS), a domain adaptation framework that freezes the feature extractor and adapts the model by optimizing only the linear decision planes.
  FPS formulates adaptation as a principled boundary search guided by \emph{explicit target-domain structural priors}, enabling effective and interpretable decision-plane refinement without explicit feature alignment or encoder fine-tuning.

  \item \textbf{Effectiveness, robustness, and scalability across architectures and domains.}
  Extensive experiments show that FPS achieves competitive performance on standard benchmarks .
  Moreover, FPS remains robust under controlled stress conditions (e.g., feature-geometry degradation, class imbalance, and label-space mismatch) and scales naturally to foundation models by avoiding expensive encoder updates.
\end{enumerate}

\section{RELATED WORKS}
\subsection*{Feature Fine-tuning }

Most UDA methods typically fine-tune the deep feature extractor to reduce the discrepancy between source and target domains 
\cite{ganin2016domain, Zhang2018CollaborativeAA, Chen2020AdversarialLearnedLF}. 
Adversarial alignment approaches further improve cross-domain feature consistency by learning domain-invariant representations, 
such as DANN \cite{ganin2016domain}, CDAN \cite{Zhang2018CollaborativeAA}, and SDAT \cite{rangwani2022closer}, 
which stabilizes adversarial optimization during feature alignment.
Other approaches improve adaptation through self-training or consistency regularization. 
For instance, SHOT \cite{Liang2020DoWR} freezes the source classifier while adapting the feature extractor via self-training. 
MIC \cite{Hoyer_2023_CVPR} enforces prediction consistency between masked and original images to enhance robustness under domain shift. 
Feature regularization methods have also been explored. 
For example, BSP \cite{chen2019transferability} penalizes dominant principal components to enhance class discriminability by promoting more homogeneous eigenvalue distributions. 
Wang and Breckon further proposed strategies for optimizing feature fine-tuning to achieve improved feature alignment performance~\cite{wang2023fine}. 
In contrast to these approaches that rely on modifying the feature extractor, FPS is motivated by the observation that, for strong encoders pre-trained at scale, the feature space often retains transferable discriminative structures across domains. 
This suggests that cross-domain performance can be improved by adjusting decision boundaries within a frozen feature space, rather than altering the feature representations themselves.

\subsection*{Unsupervised Domain Adaptation via Entropy}
A key UDA approach involves refining the classifier decision planes to reduce errors induced by target domain bias. 
A fundamental idea is entropy minimization on unlabeled target samples, which encourages the decision boundary to lie in low-density regions~\cite{NIPS2004_96f2b50b}. 
MME~\cite{saito2019semi} alternates maximizing and minimizing target conditional entropy to tighten class clusters, while MCC~\cite{jin2020minimum} penalizes ambiguous predictions that span multiple categories to alleviate category confusion.
Recent studies further highlight that entropy minimization alone may lead to degenerate solutions, motivating explicit diversity enhancement.
MEDM~\cite{9537640} systematically analyzes the trade-off between conditional entropy minimization and diversity maximization, showing that improving target discriminability requires balancing confidence with label-space coverage. 
From an information-theoretic perspective, SIDA~\cite{DBLP:journals/corr/abs-2110-12184} proposes maximizing  surrogate mutual information between intra-class features under a surrogate joint distribution, and provides an explicit upper bound that relates target risk to a negative Information Maximization (IM) term, a domain-discrepancy term, and a surrogate-bias term.
Our FPS also leverages entropy-based regularization to position decision planes in low-density regions. 
Compared with IM-style baselines, FPS operates in a frozen feature space and augments the objective with explicit structural priors including class-coverage balancing and random-pooling consistency,so that boundary refinement remains stable and interpretable under controlled perturbations.

\subsection*{Vision-Language Pretrained Models for UDA}
Large multimodal pretrained models (e.g., CLIP~\cite{Radford2021LearningTV}, BLIP~\cite{li2022blip}) provide rich, transferable embeddings that enhance UDA performance even without adaptation, and have become a focal point in recent research endeavors. HVCIP \cite{vesdapunt2024hvclip} demonstrates on the Domain-Net benchmark \cite{peng2019moment} that the zero-shot classifier of CLIP significantly outperforms previous UDA methods, highlighting the advantages of visual-linguistic pre-training. Lai et al. \cite{lai2024empowering} builds on CLIP and GLIP~\cite{li2022grounded}, and adopts a strategy of freezing their visual and textual encoders while employing just-in-time tuning with lightweight adapters tailored for UDA. Their approach involves adjusting only a task-specific prompter and a minimal classifier, while preserving the original CLIP knowledge. CLIP-Adapter \cite{gao2024clip} integrates pre-trained CLIP features into a new task head via residual concatenation, thereby bypassing the requirement for complete fine-tuning of the encoder. FPS directly simplifies the problem by completely freezing the model. This simplification also avoids the computational cost pressure associated with fine-tuning large model encoders, enabling convenient utilization of various large multimodal models.

\section{MOTIVATION AND PRELIMINARY ANALYSIS}\label{sec21}

Pretrained models such as ResNet \cite{he2016deep} and Vision Transformers (ViT) \cite{dosovitskiy2020image}, trained on large-scale datasets, exhibit strong efficacy in improving domain adaptation, particularly under source-domain sample scarcity. These models typically serve as encoders, mapping samples (e.g., images) to high-dimensional feature representations, with each sample represented as a point in the embedding space. A set of decision boundaries is then used in this feature space to classify different sample categories. 

Conventional approaches often fine-tune the encoder to achieve a refined feature space, where source and target domain distributions align more closely. This relies on two core assumptions: 
(1) performance degradation on the target domain is attributed to limitations of the encoder in capturing target-domain characteristics, leading to distributional divergence; 
(2) improving the encoder under appropriate assumptions can yield a more transferable representation.
When the source and target domains differ substantially, samples may exhibit domain-specific cues beyond category semantics, so enforcing strict alignment can distort useful representations. Moreover, fine-tuning high-capacity encoders with limited target data may be under-constrained, making the resulting feature-space changes difficult to characterize and potentially less reliable.

If the frozen encoder already provides sufficiently robust features such that the target-domain gap is dominated by decision-boundary misalignment rather than representation failure, then freezing the encoder and refining only the boundary is a viable strategy. To investigate this hypothesis, we analyze the frozen feature space using t-SNE visualizations, intra-/inter-class distance statistics, and an oracle linear-probe analysis.

\begin{figure}[h!]
\centering
\includegraphics[width=3.3in]{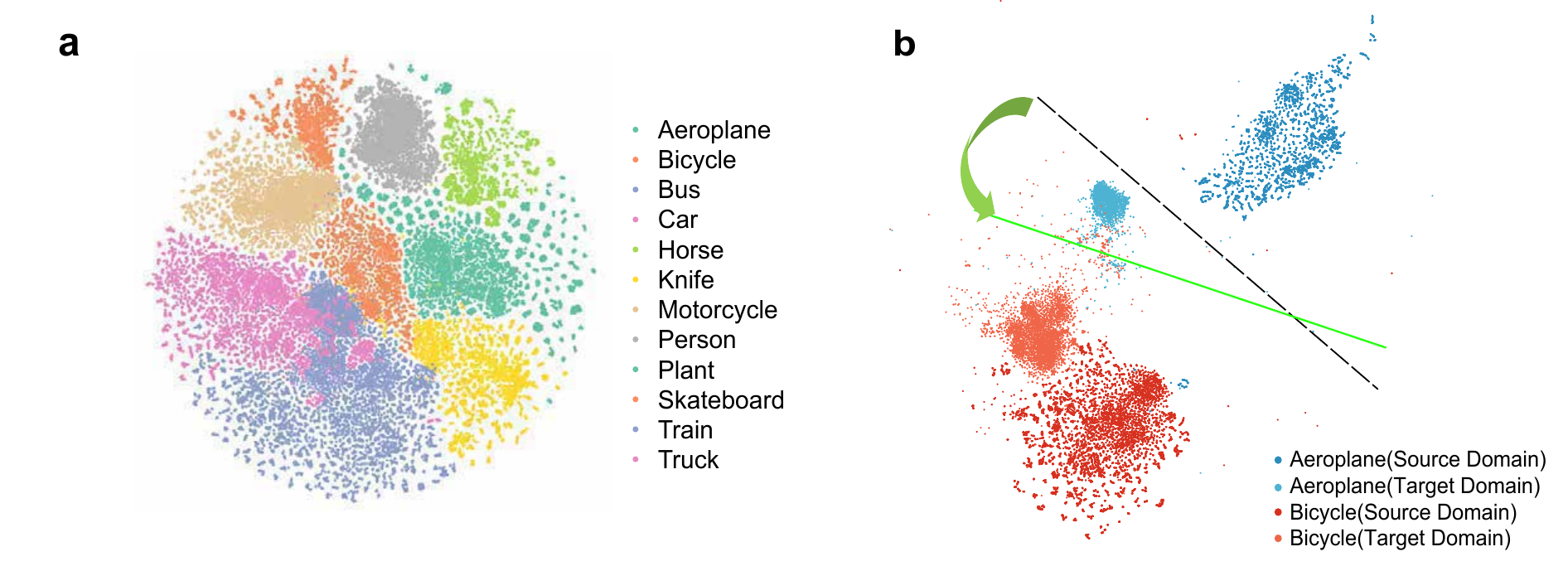}
\caption{Feature Space Visualization with Shared Decision Boundaries: (a) t-SNE dimensionality reduction of all class samples in the source domain of VisDA-2017, illustrating global feature distribution; (b) t-SNE visualization of two representative categories from both source and target domains in VisDA-2017, highlighting cross-domain alignment and boundary-sharing characteristics.}
\captionsetup{font={scriptsize}} 
\label{feature distribution pattern a}
\end{figure}
\begin{figure}[h!]
\centering
\includegraphics[width=3.3in, trim=0cm 0.0cm 0cm 0.0cm, clip]{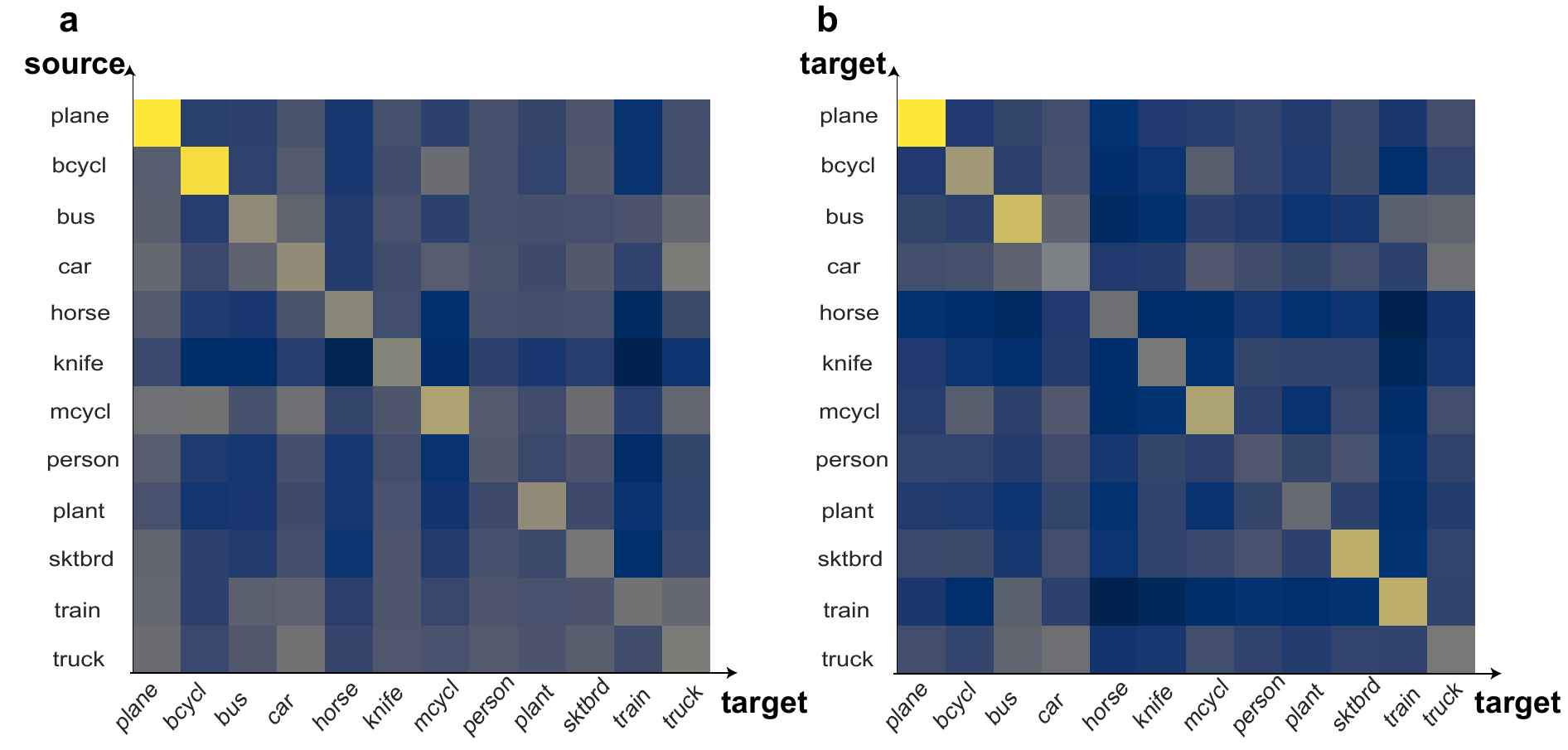}
\caption{Inter-class average distance matrix in VisDA-2017. (a) Target domain and source domain. (b) Target domain and target domain.}
\captionsetup{font={scriptsize}} 
\label{feature distribution pattern b}
\end{figure}

The t-SNE visualizations suggest a consistent geometric pattern: within a single domain, distinct classes form well-separated clusters (Fig.~\ref{feature distribution pattern a}a). Cross-domain analysis shows the class structure is largely preserved despite domain shifts, with minimal inter-class overlap across domains (Fig.~\ref{feature distribution pattern a}b). Distance metrics further support this observation: intra-class samples maintain closer proximity than inter-class samples, even under domain variations (Fig.~\ref{feature distribution pattern b}). We refer to this empirical trend as the intra-class clustering and inter-class separation pattern.

As illustrated in Fig.~\ref{feature distribution pattern a}b (black dashed line), the decline in target-domain accuracy can often be associated with a misalignment between the decision boundary learned from the source domain and the target-domain feature distribution.  As illustrated by the solid green line and green arrow in Fig. \ref{feature distribution pattern a}b, this observation suggests that cross-domain performance may be improved by adjusting the decision boundary while keeping the feature representations unchanged.

\begin{figure}[h!]
\centering
\includegraphics[width=3.5in,]{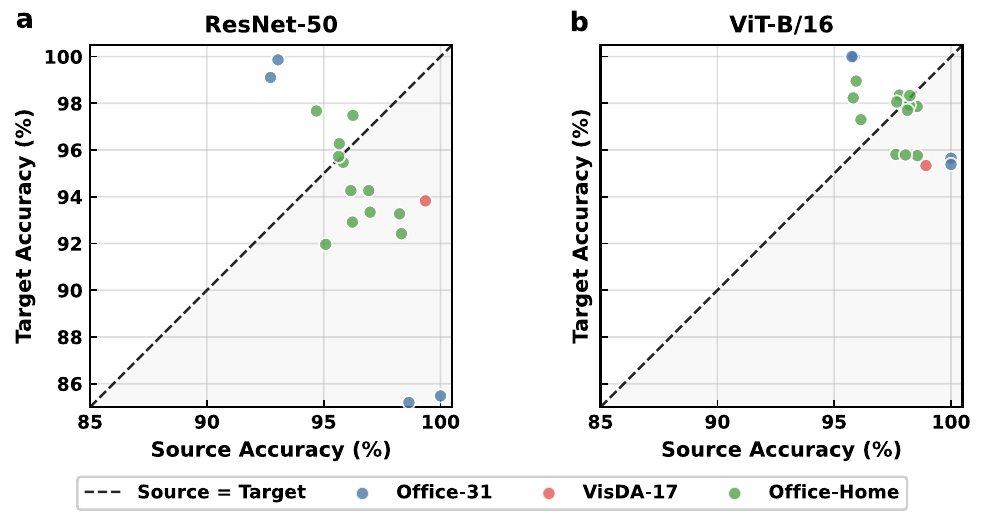}
\caption{The oracle accuracy under joint source-target domain supervision, shown for ResNet (a) and ViT-B/16 (b) across three benchmarks.}
\captionsetup{font={scriptsize}} 
\label{feature distribution pattern c}
\end{figure}

To validate this hypothesis, we froze the encoder parameters and trained linear classifiers using both source and target domain labels. As shown in Fig.~\ref{feature distribution pattern c}, the resulting classifiers achieve higher accuracy than current UDA methods on most transfer tasks (Table~\ref{performance}). This empirical result suggests that transferable decision boundaries may already exist in the pretrained feature space. Note that this observation is based on an oracle setting and is not achievable under standard UDA assumptions. It serves as an indicative analysis of the separability of the frozen feature space, suggesting that boundary refinement can be an effective strategy under appropriate conditions.

\section{METHODOLOGY}
In unsupervised domain adaptation (UDA), the labeled samples in the source domain are denoted as $\mathcal{D}_s = \{ (\mathbf{x}_{s}^i, y_{s}^i) \}_{i=1}^{N_{s}}$, where $N_{s}$ is the number of source samples. The unlabeled target domain samples are represented by $\mathcal{D}_t = \{ \mathbf{x}_{t}^i \}_{i=1}^{N_{t}}$. Traditional UDA aims to align feature representations between domains by optimizing a feature extractor. In contrast, our approach minimizes modifications to the feature extractor and instead optimizes the decision boundaries of a linear classification layer.

The classification layer is defined as:
\begin{equation}
\mathbf{y}_i = \mathbf{W}^\top \mathbf{x}_i + \mathbf{b},
\label{linear}
\end{equation}
where $\mathbf{\Theta} = \{\mathbf{W}, \mathbf{b}\}$ denotes the model parameters. The class probability vector is computed via $\mathbf{p}_i = \mathrm{Softmax}(\mathbf{y}_i)$.

\subsection{Latent Priors}\label{latents}
\begin{figure}
\centering
\includegraphics[width=3.3in, trim=0cm 1.2cm 0cm 0.7cm, clip]{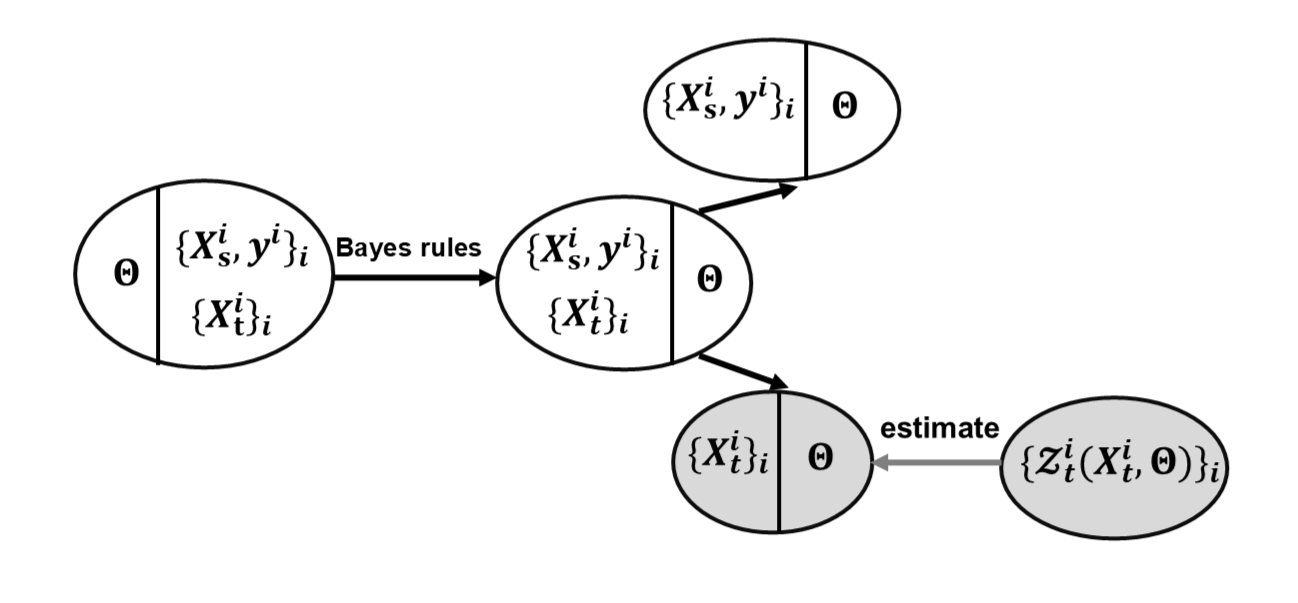}
\caption{A Bayesian Rule-Based Cross-Domain Learning Pipeline. The diagram illustrates a three-stage machine learning workflow for domain adaptation: (1) Initial State contains model parameters ($\mathbf\Theta$) and unlabeled target domain data $\{\textbf{\textit{X}}_\text{t}\}_i$ alongside labelled source domain data $\{\textbf{\textit{X}}_\text{s}^i, \textbf{\textit{\textbf{y}}}_\text{s}^i\}_i$; (2) Intermediate State reflects parameter updates via Bayesian inference, where source domain knowledge is propagated to adapt $\mathbf\Theta$ for the target domain; (3) Final Prediction generates target domain predictions $\mathcal{Z}_t^i(\textbf{\textit{X}}_\text{t}^i,\mathbf\Theta)$ after full optimization. In our study, $\mathbf\Theta$ solely represents the parameters of the final linear classification layer.
}
\captionsetup{font={scriptsize}} 
\label{fig:overall}
\end{figure}
We formulate the adaptation process as a Maximum A Posteriori (MAP) estimation problem. Given the source and target data, the optimal parameters $\mathbf{\Theta}^*$ are obtained by:
\begin{equation}
\mathbf{\Theta}^* = \arg\max_{\mathbf{\Theta}} \log p(\mathbf{\Theta} \mid \mathcal{D}_s, \mathcal{D}_t).
\label{eq:map}
\end{equation}
Using Bayes' theorem and assuming a uniform prior $p(\mathbf{\Theta})$, the objective is proportional to the joint likelihood:
\begin{align}
\log p(\mathbf{\Theta} \mid \mathcal{D}_s, \mathcal{D}_t) &\propto \log p(\mathcal{D}_s, \mathcal{D}_t \mid \mathbf{\Theta}) \notag \\
&= \underbrace{\log p(\mathcal{D}_s \mid \mathbf{\Theta})}_{\text{Supervised Term}} + \underbrace{\log p(\mathcal{D}_t \mid \mathbf{\Theta})}_{\text{Unsupervised Term}},
\label{eq:joint_likelihood}
\end{align}
where we assume conditional independence between the source and target domains given $\mathbf{\Theta}$.

While the supervised term $\log p(\mathcal{D}_s \mid \mathbf{\Theta})$ is easily optimized using standard cross-entropy loss, the unsupervised term $\log p(\mathcal{D}_t \mid \mathbf{\Theta})$ is intractable as target labels are missing. To introduce a tractable surrogate, we posit latent variables $\mathbf{Z}$ that capture underlying target-domain structures (jointly influenced by $\mathcal{D}_t$ and $\mathbf{\Theta}$). The target likelihood can then be written by marginalizing over $\mathbf{Z}$:
\begin{equation}
p(\mathcal{D}_t \mid \mathbf{\Theta}) = \int p(\mathcal{D}_t \mid \mathbf{Z}, \mathbf{\Theta})\, p(\mathbf{Z} \mid \mathbf{\Theta}) \, d\mathbf{Z}.
\end{equation}

Instead of explicitly computing the integral, we adopt a MAP-style plug-in approximation, where the dominant contribution is assumed to come from a single inferred latent configuration. Concretely, we assume that the latent variable is deterministically inferred from data and parameters via
\begin{equation}
\mathbf{Z}^*(\mathbf{\Theta}) = \mathcal{Z}(\mathcal{D}_t, \mathbf{\Theta}),
\end{equation}
which corresponds to a degenerate posterior
\begin{equation}
p(\mathbf{Z} \mid \mathcal{D}_t, \mathbf{\Theta})
=
\delta\!\big(\mathbf{Z} - \mathcal{Z}(\mathcal{D}_t, \mathbf{\Theta})\big).
\end{equation}
Under this plug-in approximation, the target-domain likelihood can be written as
\begin{equation}
p(\mathcal{D}_t \mid \mathbf{\Theta})
\approx
p(\mathcal{D}_t \mid \mathbf{Z}^*(\mathbf{\Theta}), \mathbf{\Theta})\,
p(\mathbf{Z}^*(\mathbf{\Theta}) \mid \mathbf{\Theta}),
\label{eq:approx_likelihood}
\end{equation}
where $\mathbf{Z}^*(\mathbf{\Theta}) = \mathcal{Z}(\mathcal{D}_t, \mathbf{\Theta})$. 

We treat the conditional term $p(\mathcal{D}_t \mid \mathbf{Z}^*, \mathbf{\Theta})$ as a surrogate whose $\mathbf{\Theta}$-dependence is typically weaker than the dependence of the inferred structure $\mathbf{Z}^*(\mathbf{\Theta})$ on $\mathbf{\Theta}$. Consequently, optimization is driven primarily by the structural score induced via $\mathbf{Z}^*(\mathbf{\Theta})$, which serves as a tractable, data-dependent regularizer for refining the decision boundary.
Accordingly, we approximate the target term up to $\mathbf{\Theta}$-independent constants:
\begin{equation}
p(\mathcal{D}_t \mid \mathbf{\Theta})
\propto
p(\mathbf{Z}^*(\mathbf{\Theta}) \mid \mathbf{\Theta})
\approx
p_{\mathrm{prior}}\!\big(\mathbf{Z}^*(\mathbf{\Theta})\big).
\end{equation}
Taking the logarithm yields the practical objective:
\begin{equation}
\log p(\mathcal{D}_t \mid \mathbf{\Theta})
\approx
\log p_{\mathrm{prior}}\!\big(\mathcal{Z}(\mathcal{D}_t, \mathbf{\Theta})\big).
\end{equation}

Then, the joint optimization objective for fine-tuning the decision planes becomes:
\begin{equation}
\arg\max_{\mathbf{\Theta}}
\ \log p(\mathcal{D}_s \mid \mathbf{\Theta})
\ +\
\log p_{\mathrm{prior}}\!\left(\mathcal{Z}(\mathcal{D}_t,\mathbf{\Theta})\right).
\label{eq:final_objective}
\end{equation}
 The latent variables characterize how the decision boundary parameterized by $\boldsymbol{\Theta}$ interacts with the fixed sample features. 
If such variables exhibit salient and structured distributional patterns in the target domain, their empirical distributions can be exploited as constraints of the decision-boundaries. 
Under this framework, the central objective is to identify and construct suitable latent variables $\mathbf{Z}$ that can guide principled decision-boundary adjustment; several such variables are introduced in the following subsubsections. The corresponding overall framework is shown in Fig.~\ref{fig:overall}.

The efficacy of the proposed latent structural prior can be further characterized by the generalization properties of the resulting decision boundaries. Under the PAC-Bayes framework \cite{mcallesterPACBayesianTheorems1999}, we consider the target domain risk of a Gibbs classifier $Q$, which represents a posterior distribution over the parameter space $\mathbf{\Theta}$ centered around the optimized estimate $\mathbf{\Theta}^*$. Following the change-of-measure approach for domain adaptation \cite{pmlr-v48-germain16, acunaFDomainAdversarialLearning2021}, the model is effectively constrained to regions of the parameter space that satisfy the structural requirements of the target domain.

The following theorem establishes a formal upper bound on the target risk, drawing on the theory of importance sampling in PAC-Bayes \cite{pmlr-v48-germain16}:

\begin{theorem}[Target Risk Control]
\label{thm:target_risk}
Assume the target distribution $\mathcal{D}_t$ is absolutely continuous w.r.t. the source distribution $\mathcal{D}_s$, and let $C_{t\leftarrow s} = \sqrt{1+\chi^2(\mathcal{D}_t\|\mathcal{D}_s)}$ denote the domain-gap coefficient. Let $P$ be a reference prior over $\mathbf{\Theta}$. For any $\delta\in(0,1)$, with probability at least $1-\delta$ over the draw of $\mathcal{D}_s$, the following bound holds for all posteriors $Q$:
\begin{equation}
\mathcal{R}_t(Q) \le C_{t\leftarrow s} \cdot \sqrt{ \widehat{\mathcal{R}}_s(Q) + \frac{\mathrm{KL}(Q\|P) + \log\frac{1}{\delta}}{2N_s} },
\label{eq:risk}
\end{equation}
where $\widehat{\mathcal{R}}_s(Q) = \mathbb{E}_{\mathbf{\Theta} \sim Q} [\widehat{\mathcal{R}}_s(\mathbf{\Theta})]$ is the expected empirical source risk,$N_s$ denotes the number of samples in the source domain.
\end{theorem}

Theorem~\ref{thm:target_risk} establishes a formal link between the proposed objective and the target generalisation error. Specifically, the latent-prior term $\log p_{\mathrm{prior}}(\mathcal{Z}(\mathcal{D}_t,\mathbf{\Theta}))$ serves as a data-dependent structural constraint, whose contribution is reflected in the posterior complexity term $\mathrm{KL}(Q\|P)$ in the bound;a complete proof is provided in Appendix~\ref{app:proofs}.
By discouraging parameter configurations that violate the target-domain structural patterns, optimizing Eq.~\eqref{eq:final_objective} promotes posteriors with lower effective complexity while maintaining source-domain fit. This provides a PAC-Bayes motivation for why constraining adaptation to the classifier can improve target generalization (Appendix \ref{app:pacbayes_local_prior}).

A key implication of Theorem~\ref{thm:target_risk} is that, under the same optimization constraints, restricting adaptation to the linear classification head may lead to a tighter generalization bound than fine-tuning the entire network. Since $\mathbf{\Theta}$ in our framework contains only the decision-layer parameters $\{\mathbf{W}, \mathbf{b}\}$, the PAC-Bayes complexity term $\mathrm{KL}(Q\|P)$ reflects only the deviation of the decision boundary from its reference configuration. In contrast, full-network fine-tuning enlarges the parameter space and introduces additional complexity through representation changes, which can increase $\mathrm{KL}(Q\|P)$ when empirical source fit remains comparable. Therefore, under aligned optimization settings, head-only adaptation is expected to admit a tighter bound in Theorem~\ref{thm:target_risk}. The controlled comparisons in Section~\ref{fint_abl} provide empirical evidence consistent with this implication.

\subsubsection{Sample (Shannon) Entropy}\label{unsupervised}

To leverage the observed pattern of intra-class clustering and inter-class separation in the data, we first introduce Sample (Shannon) Entropy (SE) as a latent variable. For a given sample with predicted probabilities \(\mathbf{p}(\mathbf{x}_\text{t}^i)\) across all categories, SE is computed as:
\begin{equation}
\mathrm{SE}\left(\mathbf{x}_\text{t}^i\right) = -\mathbf{p}(\mathbf{x}_\text{t}^i)^\mathrm{T}\log\mathbf{p}(\mathbf{x}_\text{t}^i).
\end{equation}
SE has been widely adopted in domain adaptation tasks, though with varying motivations. Some approaches use the predicted probability distribution under current model parameters to generate pseudo-labels for self-training \cite{wang2024dual,li2024pseudo,chen2022semi}, thereby encouraging higher confidence in assigned categories. Others minimize SE to position decision planes in low-density regions \cite{NIPS2004_96f2b50b}.

\begin{figure}[t!]
\centering
\includegraphics[width=3.3in, trim=0cm 0.40cm 0cm 0.83cm, clip]{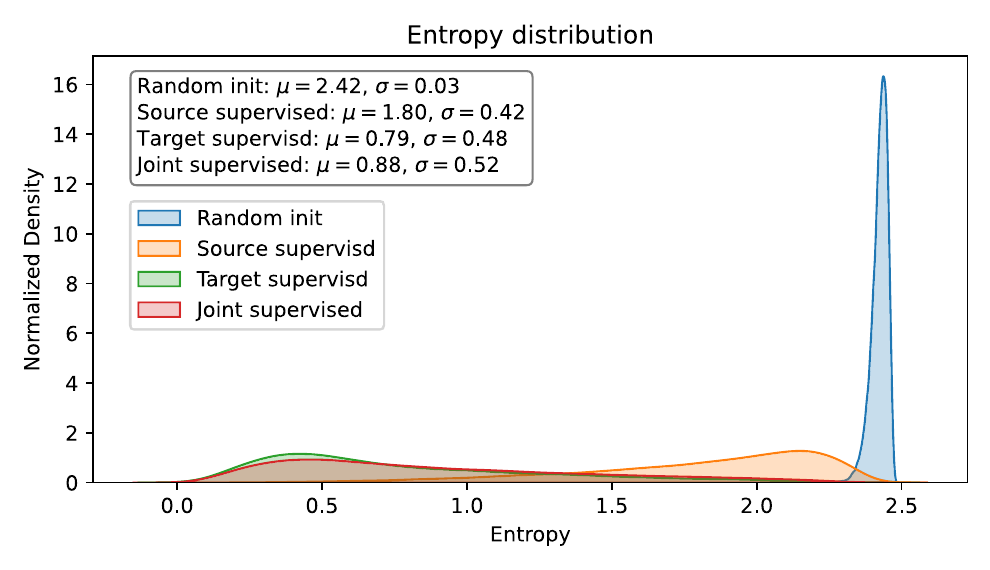}
\caption{Comparison of sample entropy distributions in the target domain of VisDA-2017. Features were extracted using a ViT-B/16 model. Distributions are visualized via kernel density estimation (KDE) with a Gaussian kernel and normalized probability density.}
\captionsetup{font={scriptsize}} 
\label{fig:entropy_display}
\end{figure}

Although we fine-tune only the final classifier, the sample entropy (SE) of target predictions provides a direct signal for adjusting decision-boundary placement in the frozen feature space. With features fixed, changes in SE can only arise from boundary movement. As shown in Fig.~\ref{fig:entropy_display}, a randomly initialised classifier yields near-uniform predictions and thus high entropy (peaking near $\log C$). With supervision, entropy decreases progressively, reaching its lowest levels under target-domain labels. If the target-supervised entropy distribution were available, it would provide a natural prior for boundary placement. In unsupervised adaptation this distribution is unavailable; however, we consistently observe that both randomly initialised and source-supervised models exhibit higher entropy than the target-supervised model. This motivates entropy minimisation as a surrogate for moving the decision boundary away from high-density regions:
\begin{equation}
\mathcal{L}_{\text{SE}} =  \mathbb{E}_{\mathbf{x}_\text{t}^i \in \mathcal{D}_{t}}[\mathrm{SE}({\mathbf{x}_\text{t}^i})]
\label{SE}
\end{equation}

Optimizing solely for SE reduction may lead to an undesirable solution in which all samples are predicted as one or a few extreme classes (Appendix Fig.~\ref{selossmap}). To prevent the decision boundaries from converging toward such a solution, we introduce the category (Shannon) entropy(CE):
\begin{equation}
\mathrm{CE}\left(\mathbf{D}_{t}\right) = -\left [   \mathbb{E}_{\mathbf{x}_{\mathrm{t}}^i \in \mathbf{D}_{t}}\mathbf{p}(\mathbf{x}_{t}^i)\right ]^\mathrm{T}\log\mathbb{E}_{\mathbf{x}_{\mathrm{t}}^i \in \mathbf{D}_{t}}\mathbf{p}(\mathbf{x}_{t}^i).
\end{equation}
We compute the mean class probabilities and then derive the Shannon entropy from these values. The corresponding category entropy loss, $\mathcal{L}_{\text{CE}}$, is defined as:
\begin{equation}
    \mathcal{L}_{\text{CE}}  = -\mathrm{CE}(\mathcal{D}_t).
\label{CE}
\end{equation}
This loss maximizes the category entropy (CE), encouraging a uniform probability distribution across all classes and preventing over concentration in a single or few categories. In entropy-based Unsupervised Domain Adaptation  methods, a similar  regularization is also employed for the same objective \cite{9537640}. Nevertheless, within the scope of this study, $\mathcal{L}_{\mathrm{CE}}$ is implemented for decision boundary optimization under the premise of frozen features.
Combining $\mathcal{L}_{\text{SE}}$ and $\mathcal{L}_{\text{CE}}$, the total entropy-related loss is:
\begin{align}
\mathcal{L}_\text{entropy}=\alpha\mathcal{L}_{\text{SE}}+(1-\alpha)\mathcal{L}_{\text{CE}}.
\label{unsup_loss}
\end{align}
Here, $\alpha$ represents the balance weight for the two losses. This forms the core loss function of our approach.
It should be emphasized that $\mathcal{L}_{\mathrm{SE}}$ and $\mathcal{L}_{\mathrm{CE}}$ respectively impose constraints on sample - level uncertainty and category - level coverage. Their combined form aligns with the Information Maximization (IM) criterion \cite{doi:https://doi.org/10.1002/047174882X.ch2,DBLP:journals/corr/abs-2110-12184,9537640}, while within a fixed feature space, this concept can be linked to geometric patterns with the prior distribution more effectively estimated.

\subsubsection{Random Pooling Constraint}\label{random_pooling}
Beyond entropy-based objectives, we introduce a latent perturbation variable tied to the pooling mechanism.In modern classifiers, global average pooling (CNNs) or the CLS token (ViTs) aggregates patch- or region-level features into a single representation; this operation can be interpreted as a weighted average over spatial tokens. We impose the prior that predictions are invariant to mild perturbations of the aggregation weights. To formalise this prior, we define \emph{random pooling}: given token features $\{\mathbf{x}_p\}_{p=1}^{N_p}$ and nonnegative random weights $\{\omega_p\}_{p=1}^{N_p}$, we compute
\begin{equation}
\mathbf{x}(\omega) =
\frac{\sum_{p=1}^{N_p}\omega_p \mathbf{x}_p}{\sum_{p=1}^{N_p}\omega_p},
\end{equation}
where $N_p$ is the number of spatial tokens (ViT patches excluding CLS, or flattened CNN feature-map locations). For CNN backbones, we sample $\omega_p\sim\mathcal{U}(0,1)$. For ViT backbones, we set $\omega_p=\eta_p^2$ with $\eta_p\sim\mathcal{U}(0,1)$ to increase weight variance while ensuring $\omega_p\ge 0$. 

\begin{figure}
\centering
\includegraphics[width=3.3in, trim=0cm 0.45cm 0cm 0.83cm, clip]
{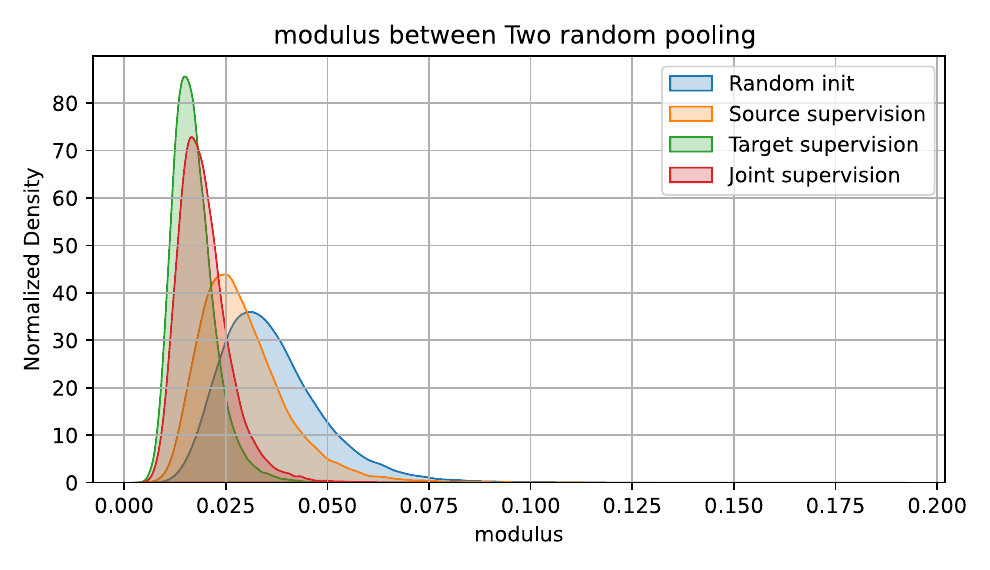}
\caption{$\mathrm{CR}$ distributions in the target domain from the VisDA-2017 dataset under different supervision. Other settings remain consistent with those described in Fig.~\ref{fig:entropy_display}.  $\mathrm{CR}$ is normalized by $(\|{\mathbf{y}}_t^i\|_2+\|\hat{\mathbf{y}}_t^i\|_2)/2$.}
\captionsetup{font={scriptsize}} 
\label{fig:randompool}
\end{figure}

Two random pooling operations on $\mathbf{x}_{p}$ produce features $\mathbf{x}_\text{t}^i$ and $\hat{\mathbf{x}}_\text{t}^i$, leading to predictions $\mathbf{y}_\text{t}^i$ and $\hat{\mathbf{y}}_\text{t}^i$. The prediction discrepancy (CR) between perturbations is:
\begin{align}
    \text{CR}(\mathbf{y}_\text{t}^i, \hat{\mathbf{y}}_\text{t}^i) = \|{\mathbf{y}}_\text{t}^i - \hat{\mathbf{y}}_\text{t}^i\|_2.
\end{align}
The name "CR" is derived from the abbreviation of the Consistency Regularization term. To validate that perturbations minimally affect classification, we analyze the CR distribution across supervision regimes (Fig.~\ref{fig:randompool}). When the supervised data approximates the target domain, the CR distribution converges to zero. This implies that robust decision boundaries reduce sensitivity to pooling variations, ensuring consistent classification for independently pooled features \cite{gholamalinezhad2020pooling}.
Given these distributional differences, CR serves as a latent variable for optimizing decision boundaries. To circumvent estimating CR's prior distribution, we define a loss function to minimize CR:
\begin{align}
\mathcal{L}_\mathrm{CR} & = \frac{\mathbb{E}_{x \in \mathbf{X}_\text{t} } \left[ \|{\mathbf{y}}_\text{t}^i - \hat{\mathbf{y}}_\text{t}^i\|_2 \right]}{C},
\end{align}
where $C$ is the number of classes. Unlike the entropy loss describing the relationship between the decision boundary and feature distribution, CR primarily serves as a latent variable describing the consistency of the decision boundary under data perturbations. $\mathcal{L}_{\mathrm{CR}}$ provides a fine-grained structural constraint that complements entropy-based terms and stabilises boundary search under controlled perturbations. Notably, the random pooling is applied on-the-fly to the cached tokens to produce $\mathbf{x}_\text{t}^i$ and $\hat{\mathbf{x}_\text{t}^i}$ without re-running the encoder. 

\subsection{Additional Detail}
The two latent variables, SE($\cdot$) and CR($\cdot$), introduced earlier form the core of our decision boundary fine-tuning. Building upon this foundation, we will now introduce additional details that collectively constitute the entire algorithm.

\subsubsection{Dynamically Weighted Loss Function}\label{class_weight}
Currently, we have the supervised cross-entropy loss ($\mathcal{L}_\text{SCE}$) for labeled source domain data, along with $\mathcal{L}_\text{entropy}=\alpha\mathcal{L}_{\text{SE}}+(1-\alpha)\mathcal{L}_{\text{CE}}$ and $\mathcal{L}_\text{CR}$ for unlabeled target domain data, all employed throughout the training process. 
Through experimentation, we observed that prioritizing $\mathcal{L}_{\mathrm{SCE}}$  initially, followed by gradually increasing the weights of $\mathcal{L}_{\mathrm{SE}}$, $\mathcal{L}_{\mathrm{CE}}$ and $\mathcal{L}_{\mathrm{CR}}$, yields optimal results in most tasks. Specifically, focusing primarily on the categorical entropy loss ($\mathcal{L}_{\mathrm{CE}}$) at the outset, followed by progressively increasing the weight of the sample entropy loss ($\mathcal{L}_{\mathrm{SE}}$), is advantageous. Therefore, we have designed a dynamically weighted loss function defined as follows:
\begin{align}
    \mathcal{L} &= \beta_t\mathcal{L}_{\text{SCE}} + (1-\beta_t)\left[\mathcal{L}_{\text{entropy}}\right. \\ \notag
    &\left. + \lambda\mathcal{L}_{\mathrm{CR}} \right],
\label{loss}
\end{align}
where, $\mathcal{L}_\text{entropy}=\alpha_t\mathcal{L}_{\text{SE}}+(1-\alpha_t)\mathcal{L}_{\text{CE}}$ and the hyperparameters $\alpha_t$, $\beta_t$ and $\lambda$ balance the loss weights and vary with the optimizing step. The hyperparameters are defined as follows:
\begin{equation}
    \alpha_t = \alpha + (\alpha_0-\alpha)\mathrm{e}^{-\frac{t}{T}},
\end{equation}
\begin{equation}
    \beta_t = \beta_0 + (\beta-\beta_0)\mathrm{e}^{-\frac{t}{T}},
\end{equation}
where, for most cases, $\alpha_0$, $\beta_0$ ,$T$ and $\lambda$ are fixed at 0.1, 0.1,1000,0.55 respectively. $\alpha$ is a hyperparameter that requires tuning. $\beta$ denotes the initial weight on $\mathcal{L}_{\text{SCE}}$ and $\beta_0$ is the terminal weight, so that $\beta_t$ decays from $\beta$ to $\beta_0$ as $t$ increases. Later, we will introduce \hyperref[Intra-class distance metric]{\textit{\nameref{Intra-class distance metric}}}, an indicator used to select better hyperparameters in an unsupervised setting. Examples of hyperparameter determination in specific applications can be found in \hyperref[hyper]{\textit{\nameref{hyper}}}.

\subsubsection{Density Weighting Strategy}
Furthermore, to alleviate performance degradation caused by class imbalance, we introduce an inverse density weighting strategy. 
This approach operates on the assumption that majority classes typically form denser clusters in the feature space compared to minority classes. 
Consequently, we penalize samples with high local density by assigning them lower weights. 
The density-aware weight $w_i$ for the $i$-th sample is computed as:
\begin{equation}
    w_i = \frac{1}{ \sum_{j=1}^{N_\mathrm t} \exp({A \cdot \mathcal{S}(\mathbf{x}_i, \mathbf{x}_j)})}
\end{equation}
where $\mathcal{S}(\mathbf{x}_i, \mathbf{x}_j) = \frac{\mathbf{x}_i^\mathrm{T} \mathbf{x}_j}{\|\mathbf{x}_i\|_2 \|\mathbf{x}_j\|_2}$ represents the cosine similarity between samples $\mathbf{x}_i$ and $\mathbf{x}_j$, and $A$ (typically set to 5) serves as a scaling factor to modulate the sharpness of the density kernel. 
To ensure the weights are purely relative and preserve the total loss magnitude (i.e., maintaining an average weight of 1), we apply the following normalization:
\begin{equation}
    \tilde{w}_i = \frac{N_\mathrm{t}}{\sum_{k=1}^{N_\mathrm{t}} w_k} w_i.
    \label{eq:weight}
\end{equation}
These calibrated weights are subsequently incorporated into $\mathcal{L}_{\text{SE}}$, $\mathcal{L}_{\text{CE}}$, and $\mathcal{L}_{\text{CR}}$ for the target domain. 

\subsubsection{Margin-Aware Confidence Weighting}\label{Margin-Aware Sample Weighting}

When class separability is low, the decision boundary may pass through regions containing highly ambiguous target samples. In such cases, entropy minimization alone may be less reliable, since predictions near the boundary tend to be uncertain and may disproportionately influence the regularization process \cite{grandvalet2004entropy,arazo2020pseudo}. To mitigate this issue, we reduce the influence of low-distinctiveness samples by assigning confidence-aware weights to target instances according to their normalized prediction margins. This design is motivated by the geometric interpretation of margins in multiclass classifiers \cite{crammer2001multiclass}.

For a target sample, let $y_i^{(1)}$ and $y_i^{(2)}$ denote the largest and second-largest logits, and let $w^{(1)}$ and $w^{(2)}$ be the classifier weight vectors associated with the corresponding classes. We define the normalized logit gap as
\begin{equation}
m_i =
\frac{y_i^{(1)} - y_i^{(2)}}
{\left\| w^{(1)} - w^{(2)} \right\|_2}.
\end{equation}
This normalization reflects the geometric separation between the two most competitive classes in the classifier space, so that the resulting score captures both prediction confidence and the local stability of the decision boundary.

To convert the margin into a soft confidence weight, we first estimate a margin threshold $\theta$ from the empirical quantile of target margins,
\begin{equation}
\theta = \operatorname{Quantile}(\{m_i\}, q),
\end{equation}
and then apply a sigmoid mapping:
\begin{equation}
\omega'_i =
\sigma\!\left(
\frac{m_i - \theta}{\tau} - b
\right),
\label{eq:margin}
\end{equation}
where $\sigma(\cdot)$ denotes the sigmoid function, $\tau$ controls the transition smoothness, and $b$ is a bias term used to calibrate the weight at the threshold. Specifically, given a desired boundary weight $\rho \in (0, 0.5)$, we set
\begin{equation}
b = \log \frac{1-\rho}{\rho}.
\end{equation}

Under this formulation, samples with larger normalized margins receive higher weights, while ambiguous samples near the decision boundary are automatically down-weighted. The resulting confidence-aware weighting modulates the target regularization term, allowing the optimization to focus more on reliable target samples, which helps improve robustness under substantial domain shift. Similar confidence-based filtering and sample selection strategies have been widely adopted in semi-supervised learning and pseudo-labeling methods \cite{arazo2020pseudo,sohn2020fixmatch}.

In our implementation, we adopt the normalized\_logit\_gap criterion together with the quantile\_sigmoid mapping, with hyperparameters $q=0.15$, $\tau=0.2$, and $\rho=0.2$. These values are fixed across all experiments unless otherwise specified.

\subsubsection{Intra-Class Distance Metric} \label{Intra-class distance metric}    
Similar to methods such as k-means, if samples from the same category exhibit clustering tendencies, the optimal decision boundary should minimize intra-category sample distances. If this assumption holds, it allows us to assess decision boundary optimization performance and select optimal hyperparameter configurations without relying on labeled target-domain data.

Building on these insights, we propose a non-parametric metric, the Intra-Class Distance Metric (ICDM). The centroid of class $c$ in the target domain is computed as:
\begin{equation}
\mathbf{\mu}_c = \mathbb{E}_{\mathbf{x}^i \in \mathcal{D}_c}[\mathbf{x}^i], \quad \text{where } \mathcal{D}_c = \{\mathbf{x}^i \in \mathbf{D}_\text{t} \mid \hat{y}_i = c\}.
\label{eq:centroid}
\end{equation}
Here, $\hat{y}_i$ denotes the model's predicted pseudo-labels. Intra-class similarity is measured as the squared Euclidean distance between a sample and its corresponding centroid:
\begin{align}
   &\hat{D}_{\text{intra}} = \sqrt{\mathbb{E}_{x_i \in \mathbf{X}_\text{t}}[||\mathbf{x}^i - \mathbf{\mu}_{\hat{y}_i}||^2]}.
    \label{eq:intra_distance}
\end{align}
For cross-task visualization only, $\hat{D}_{\text{intra}}$ is normalized by the corresponding metric $D_{\text{intra}}$ computed from true labels:
\begin{equation}
    R = \frac{\hat{D}_{\text{intra}}}{D_{\text{intra}}}.
\end{equation}
Here, $D_{\text{intra}}$ is solely used to normalize $\hat{D}_{\text{intra}}$ across tasks, facilitating comparison, but is not required for parameter selection.

\begin{figure}[h!]
\centering
\includegraphics[width=3.3in, trim=0cm 0cm 0cm 0.cm, clip]{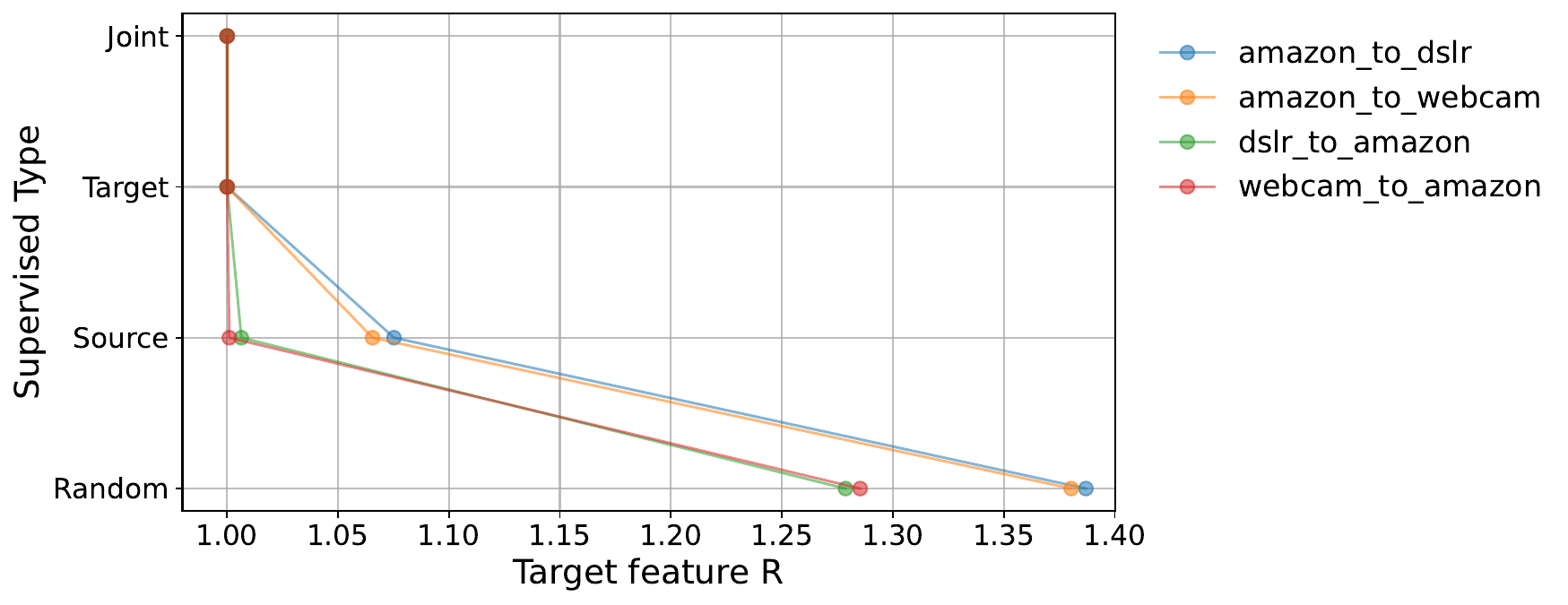}
\caption{Comparison of normalized intra-class distance (R) across different transfer tasks from the Office-31 dataset under different supervision. The x-axis represents the target feature R , while the y-axis lists various supervision types, including Random, Source, Target, and Joint strategies. Each line corresponds to a distinct domain adaptation task, with markers indicating the R values under different supervision settings. }
\captionsetup{font={scriptsize}} 
\label{fig:R}
\end{figure}

As shown in Figure~\ref{fig:R}, $R$ varies across tasks depending on supervision levels. Notably, when incorporating the source domain data, $R$ significantly decreases compared to random class partitioning; furthermore, $R$ declines even more after introducing the target domain data. This indicates that as the decision planes become more aligned with the target domain data, $R$ generally decreases.

\subsubsection{Feature Preprocessing}\label{train flow}
We precompute and store all feature vectors of the dataset, thereby fully leveraging the statistical information of the target domain. To maintain consistency in search strategies across different backbones, the features undergo standardization (zero-mean, with the standard deviation normalized to 2.5):
\begin{equation}
\mathbf{x}' = \frac{\mathbf{x} - \mu}{\sigma} \cdot s
\end{equation}
where 
\begin{align*}
\mu &= \mathbb{E}_{\mathbf{x}_{\mathrm{t}}^i \in \mathbf{D}_s \cup \mathcal{D}_t} \left[ \mathbf{x}_{t}^i \right], \\
\sigma &= \sqrt{\mathbb{E}_{\mathbf{x}_{\mathrm{t}}^i \in \mathbf{D}_s \cup \mathcal{D}_t} \left[ (\mathbf{x}_{t}^i - \mu)^2 \right]}.
\end{align*} 
and $s$ is set to the target standard deviation of 2.5. Note that for ResNet models, the final feature output (after ReLU and pooling layers, shown in Appendix Fig.\ref{app: add Feat Anal}a) deviates more from a normal distribution compared to ViT-based models (Appendix Fig.\ref{app: add Feat Anal}b). Given the non-negativity of features, we take the square root of the original features before distribution transformation to make them more closely resemble a normal distribution (Appendix Fig.\ref{app: add Feat Anal}c).

\subsubsection{Overall Process}
The overall training process is detailed in Algorithm \ref{alg:fps}. We employ the SGD optimizer \cite{L2010Large} with momentum 0.9 to iteratively adjust the decision boundary for a fixed number of optimization steps  (36,000 in our experiments). 
A linearly increasing learning rate schedule is set as 
\[
\text{LR} = \text{base LR} \times \text{current step}.
\]
\begin{figure*}[htbp!]
\centering
\includegraphics[width=6.0in, trim=0cm 0.1cm 0cm 0.5cm, clip]{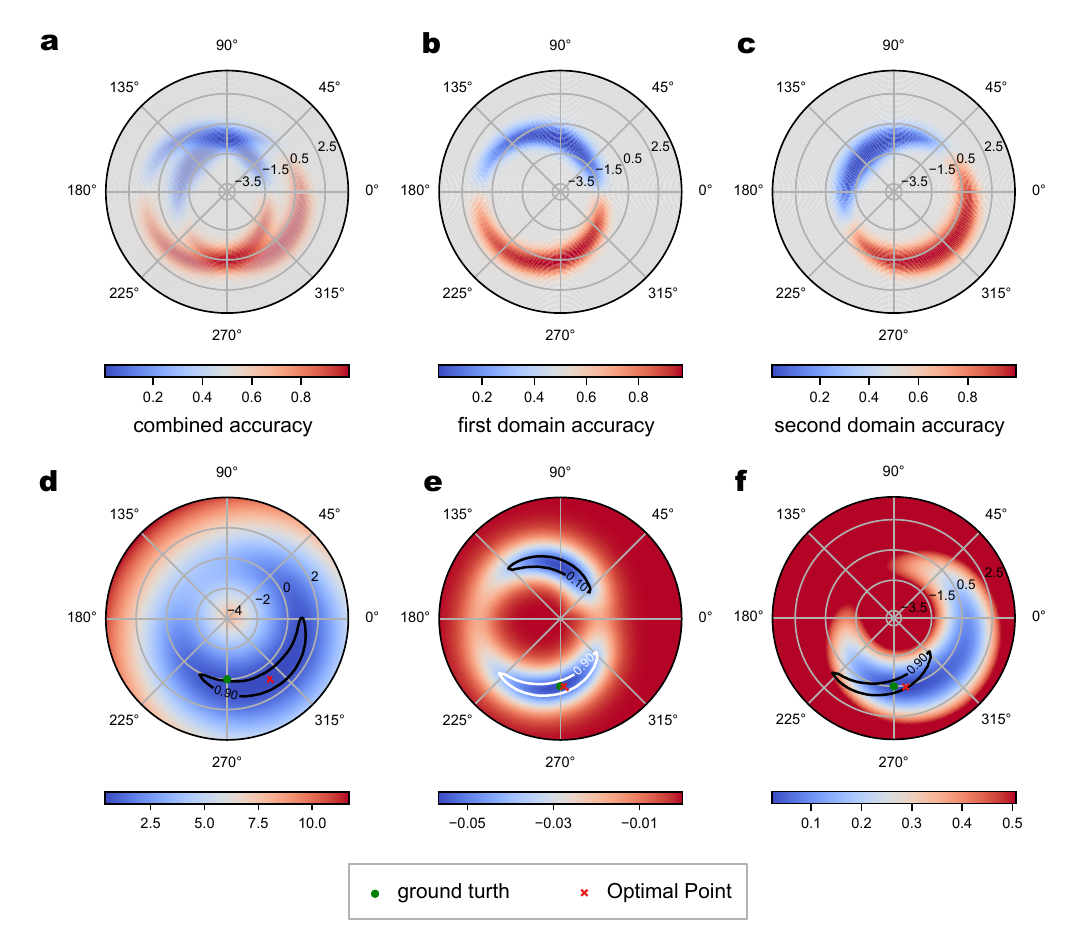}
\caption{The figures (d/e/f and g/h/i) illustrate the impact of decision boundary parameters on cross-domain classification accuracy and loss dynamics, with polar-coordinate visualizations of boundary thresholds and contour plots depicting performance trade-offs between supervised, unsupervised, and joint objectives.(d/e/f): Accuracy rates under different decision planes for the combined, target and source domains, respectively. Polar coordinates are used, where the radius represents $b$ and the angle corresponds to $\theta$. (g/h/i): Distributions of supervised, unsupervised, and joint losses under various decision boundary parameters. The black or white solid lines represent accuracy contours for the source, target, and target domains, respectively.}
\captionsetup{font={scriptsize}} 
\label{demo b}
\end{figure*}
\begin{algorithm}[htbp]
\caption{Feature Plane Searcher (FPS) Training Framework}
\label{alg:fps}
\begin{algorithmic}[1]

\Require Source features $\mathbf{X}_s$, target features $\mathbf{X}_t$; hyperparameters $A, s, q, \tau, \rho$
\Ensure Optimal classifier parameters $\theta$

\Statex
\Statex \textbf{Phase 1: Feature Preprocessing}
\State Compute global statistics $\mu,\sigma$ over $\mathbf{X}=\mathbf{X}_s\cup\mathbf{X}_t$
\State Standardize features: $\mathbf{x}'=\frac{\mathbf{x}-\mu}{\sigma}\cdot s$

\Statex
\Statex \textbf{Phase 2: Sample Weight Computation}
\State Compute similarity-based weights $\tilde{w}_i$ using Eq.~(\ref{eq:weight})

\Statex
\Statex \textbf{Phase 3: Network Training}
\For{$t = 1$ to $T$}
\State Compute margin-aware confidence weights $\omega'_i$ using Eq.~(\ref{eq:margin})
\State $\hat{w}_i \gets \tilde{w}_i \cdot \omega'_i$
    \State $\mathcal{L}_{total}=\sum_i \beta_t\mathcal{L}_{SCE}+(1-\beta_t)\tilde{w}_i\left[\mathcal{L}_{entropy}+\lambda\mathcal{L}_{CR}\right]$
    \State Update $\theta$
\EndFor

\end{algorithmic}
\end{algorithm}

\section{EXPERIMENTS}\label{EXPERIMENTS}

\subsection{Illustration Demo}\label{id}

\begin{figure}[h!]
\centering
\includegraphics[width=3.3in, trim=0cm 0.50cm 0cm 0.05cm, clip]{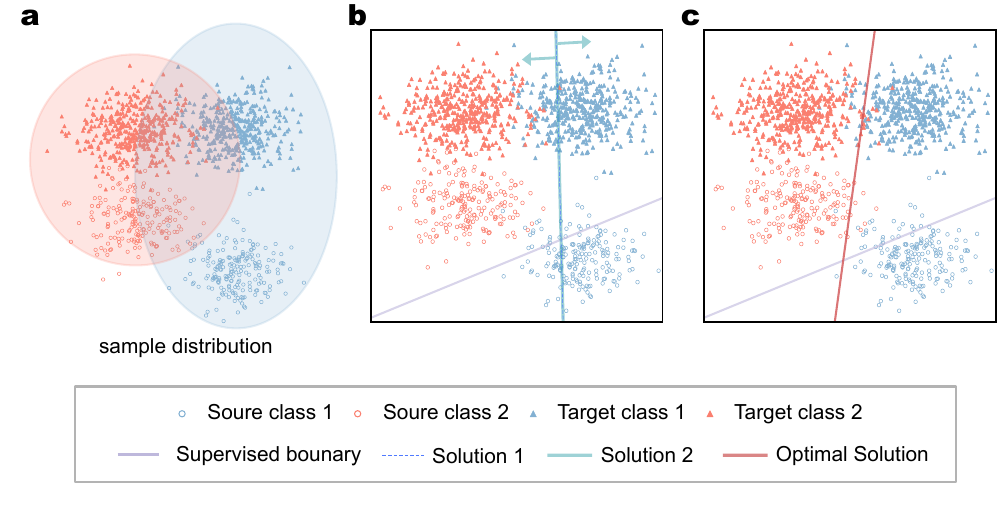}
\caption{A schematic illustrating the optimization of the decision boundary in a two-dimensional feature space. (a) Distribution of sample features. (b) Similar to (a), but with decision boundary: the solid purple line represents the supervised decision boundary of the source domain, the blue dashed line depicts the unsupervised decision boundary 1 for target domain samples, and the light blue solid line shows the unsupervised decision boundary 2 for the target domain. (c) Similar to (b), but the unsupervised decision boundary are replaced by the decision boundary derived from FPS, indicated by the solid red line.}
\captionsetup{font={scriptsize}} 
\label{demo a}
\end{figure}

Before formally conducting the experiments, we use a simplified two-dimensional scenario to illustrate our objective and optimization behavior. As shown in Fig.~\ref{demo a}, in the two-dimensional feature space, samples from the source domain and the target domain are distributed, represented by circles and triangles, respectively. These samples belong to two distinct classes, with the corresponding categories denoted by red and blue. The decision boundary is represented as:
\begin{equation}
    \sin (\theta)  x_1 + \cos (\theta)  x_2 + b = 0
\end{equation}
where \(\theta\) represents the orientation of the decision boundary, \(b\) denotes the signed offset (i.e., the distance to the origin along the normal direction), $x_1$ and $x_2$ represent the two components of the feature vector.

In Fig.~\ref{demo b}a/b/c, the accuracy under different $(\theta, b)$ is shown for the combined domain, the target domain and the source domain, respectively. Although there are differences in the distributions of the two domains, there exist decision-plane configurations that can simultaneously satisfy the accuracy requirements of both domains (Fig.~\ref{demo b}a/d). Applying $\mathcal{L}_{\text{SE}}$ and $\mathcal{L}_{\text{CE}}$ in the demo, the desired decision boundary is one of the two local minima (Fig.~\ref{demo b}e). Added with the supervised cross-entropy loss $\mathcal{L}_{\text{SCE}}$ of the labeled source domain, the decision boundary is further disambiguated and refined, converging to a unique optimum (Fig.~\ref{demo b}f). 

\subsection{Benchmark Performance}
To comprehensively validate the effectiveness of FPS, we conducted extensive experiments across three standard domain adaptation benchmarks: Office-31  \cite{2010Adapting}, Office-Home \cite{2017Deep}, and VisDA-2017 \cite{Peng2017VisDATV}. \noindent\textbf{Office-31} contains 4,600 images across 31 classes from three domains: Amazon (\textbf{A}), DSLR (\textbf{D}), and Webcam (\textbf{W}). 
\textbf{Office-Home} consists of 15,500 images covering 65 categories in four domains: Art (\textbf{Ar}), Clipart (\textbf{Cl}), Product (\textbf{Pr}), and Real World (\textbf{Rw}). 
\textbf{VisDA-2017} is a large-scale synthetic-to-real benchmark with 12 classes, comprising 152k synthetic images and 55k real images sourced from MSCOCO~\cite{Lin2014MicrosoftCC}. Our evaluation protocol follows rigorous standards by comparing against 12 representative UDA baselines, including several recent state-of-the-art methods, under identical backbone architectures (ResNet-50/101 and ViT-B/16). 

\subsubsection{Hyperparameters Settings}\label{hyper}

\begin{figure}[h!]
\centering
\includegraphics[width=3.3in,trim=0cm 0.cm 0.cm 0.cm, clip]{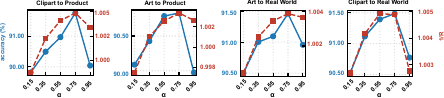}
\caption{Across several tasks in Office-Home(with ViT as the backbone), accuracy (Acc.) and the reciprocal of intra-class distance ratio $\frac{1}{R}$ (refer to the definition of R in \hyperref[Intra-class distance metric]{\textit{\nameref{Intra-class distance metric}}}) vary with respect to different values of $\alpha$.}
\captionsetup{font={scriptsize}} 
\label{R and performance a}
\end{figure}

As illustrated in Fig. \ref{R and performance a}, suitable parameter settings can be obtained by minimizing the \hyperref[Intra-class distance metric]{\textit{\nameref{Intra-class distance metric}}} without access to ground-truth labels in the target domain.
The specific settings of $\alpha,\alpha_0,\beta,\beta_0$ and base LR settings for different tasks are detailed in Tables \ref{tab:hyperparameters-final} in the Appendix \ref{app:hypersetting-benchmark}. In the image preprocessing of VisDA-2017, due to the significant variation in image aspect ratios, we first pad the data into squares, then resize them to the model's preset dimensions, and ignore the patches or tokens from the padded regions during pooling. A similar preprocessing strategy is applied to Office-Home and Office-31 when using the ViT backbone.

\subsubsection{Main Results and Analysis}

As shown in Table~\ref{performance}, FPS achieves competitive performance across multiple domain adaptation benchmarks under both convolutional and transformer-based backbones. 

On Office-Home with ResNet, FPS attains an average accuracy of 73.1\%, which is comparable to recent methods such as GSDE, while achieving the best performance on several individual tasks. Under the ViT backbone, FPS achieves an average accuracy of 86.4\%, which is competitive with methods such as MIC. On the more challenging VisDA-2017 benchmark, FPS achieves competitive results under both ResNet and ViT backbones. While methods such as CDCL and MIC achieve higher overall accuracy, FPS performs favorably on several difficult categories (e.g., truck and person for the ResNet backbone). On Office-31, FPS achieves strong and consistent performance across all tasks, obtaining the best average accuracy under both ResNet and ViT backbones. 

Overall, these results indicate that FPS provides a competitive alternative to existing methods, particularly under strong pretrained feature representations. 
At the same time, some methods that incorporate additional representation-level constraints (e.g., MIC) can yield further gains in some settings, suggesting that boundary refinement and feature adaptation may play complementary roles rather than mutually exclusive.


\begin{table*}[t!]
\caption{Accuracy (\%) on benchmarks for domain adaptation. O. H. for Office-Home, V. 17 for VisDA-17, O. 31 for Office-31. The best performance is highlighted in bold. CDTrans* employs the DeiT-base backbone \citep{Touvron2020TrainingDI}. For the Office-31, the easiest domain adaptation between Dslr (D) and Webcam (W) was not presented \citep{he2016deep}. The numbers 1 to 12 represent distinct tasks within each dataset. For Office-Home, they correspond to Ar→Cl, Ar→Pr, Ar→Rw, Cl→Ar, Cl→Pr, Cl→Rw, Pr→Ar, Pr→Cl, Pr→Rw, Rw→Ar, Rw→Cl, and Rw→Pr, respectively. For VisDA-17, they correspond to plane, bicycle, bus, car, house, knife, motorcycle, person, plant, skateboard, train, and truck, respectively. For Office-31, the first four of them correspond to A→W, W→A, A→D, and D→A, respectively. The gray backgrounds indicate the ResNet backbones (ResNet50 for  O. H./O. 31 and ResNet101 for V. 17), while the white backgrounds indicate the ViT-B/16 backbone.}
\label{performance}
\small 
\setlength\tabcolsep{4pt}
\centering
\begin{tabular}{c>{\centering}p{2.5cm}ccccccccccccc}
\toprule
\textbf{Dataset} & \textbf{Method} & \textbf{1} & \textbf{2} & \textbf{3} & \textbf{4} & \textbf{5} & \textbf{6} & \textbf{7} & \textbf{8} & \textbf{9} & \textbf{10} & \textbf{11} & \textbf{12} & \textbf{Avg.} \\
\midrule
\rowcolor{gray!20}
 & source \citep{he2016deep} & 44.0 & 65.9 & 74.0 & 52.1 & 61.0 & 65.4 & 52.2 & 40.9 & 72.9 & 64.1 & 45.3 & 77.9 & 59.6 \\
\rowcolor{gray!20}
& CDAN \citep{Zhang2018CollaborativeAA} & 50.7 & 70.6 & 76.0 & 57.6 & 70.0 & 70.0 & 57.4 & 50.9 & 77.3 & 70.9 & 56.7 & 81.6 & 65.8 \\
\rowcolor{gray!20}
& MCC \citep{jin2020minimum} & 55.1 & 75.2 & 79.5 & 63.3 & 73.2 & 75.8 & 66.1 & 52.1 & 76.9 & 73.8 & 58.4 & 83.6 & 69.4 \\
\rowcolor{gray!20}
& DALN \citep{chen2022reusing} & 57.8 & 79.9 & 82.0 & 66.3 & 76.2 & 77.2 & 66.7 & 55.5 & 81.3 & 73.5 & 60.4 & 85.3 & 71.8 \\
\rowcolor{gray!20}
& MEDM \citep{9537640} & 57.1 & 76.1 & 78.4 & 63.3 & 76.8 & 78.9 & 63.5 & 54.8 & 79.9 & 66.2 & 55.0 & 83.9 & 69.5 \\
\rowcolor{gray!20}
& SIDA \citep{DBLP:journals/corr/abs-2110-12184} & 57.2 & 79.1 & 81.7 & 67.1 & 74.5 & 77.3 & 67.2 & 53.9 & 82.5 & 71.4 & 58.7 & 83.3 & 71.2 \\
\rowcolor{gray!20}
& GSDE \citep{westfechtel2024gradual} & 57.8 & 80.2 & 81.9 & \textbf{71.3} & 78.9 & 80.5 & 67.4 & \textbf{57.2} & 84.0 & \textbf{76.1} & 62.5 & 85.7 & \textbf{73.6} \\
\rowcolor{gray!20}
\textbf{O. H.}& SDAT \citep{rangwani2022closer} & \textbf{58.2} & 77.1 & 82.2 & 66.3 & 77.6 & 76.8 & 63.3 & 57.0 & 82.2 & 74.9 & \textbf{64.7} & \textbf{86.0} & 72.2 \\
\rowcolor{gray!20}
& FPS & 55.1 & \textbf{81.8} & \textbf{83.6} & 68.3 & \textbf{82.6} & \textbf{82.3} & \textbf{68.8} & 52.8 & \textbf{84.1} & 72.6 & 59.3 & 85.3 & 73.1 \\

\hline
& source \citep{dosovitskiy2020image} & 66.2 & 84.3 & 86.6 & 77.9 & 83.3 & 84.3 & 76.0 & 62.7 & 88.7 & 80.1 & 66.2 & 88.7 & 78.7 \\
& CDTrans* \citep{xu2021cdtrans} & 68.8 & 85.0 & 86.9 & 81.5 & 87.1 & 87.3 & 79.6 & 63.3 & 88.2 & 82.0 & 68.0 & 90.6 & 80.5 \\
& TVT \citep{yang2023tvt} & 74.9 & 86.8 & 89.5 & 82.8 & 88.0 & 88.3 & 79.8 & 71.9 & 90.1 & 85.5 & 74.6 & 90.6 & 83.6 \\
& SSRT \citep{sun2022safe} & 75.1 & 88.9 & 91.0 & 85.1 & 88.2 & 89.9 & \textbf{85.0}& 74.2 & 91.2 & 85.7 & 78.5 & 91.7 & 85.4 \\
& SDAT \citep{rangwani2022closer} & 70.8& 87.0 & 90.5 & 85.2 & 87.3 & 89.7 & 84.1 & 70.7 & 90.6 & 88.3 & 75.5 & 92.1 & 84.3 \\
& MIC\cite{Hoyer_2023_CVPR} & \textbf{80.2} & 87.3 & 91.1 & \textbf{87.2} & 90.0 & 90.1 & 83.4 & 75.6 & 91.2 & \textbf{88.6} & \textbf{78.7} & 91.4 & 86.2 \\
& FPS & 78.2 & \textbf{91.3} & \textbf{91.4} & 84.7 & \textbf{91.7} & \textbf{91.5} & 84.4 &\textbf{75.9} & \textbf{91.5} & 84.8 & \textbf{78.7} & \textbf{92.7} & \textbf{86.4} \\

\midrule
\rowcolor{gray!20}
 & source \citep{he2016deep} & 55.1 & 53.3 & 61.9 & 59.1 & 80.6 & 17.9 & 79.7 & 31.2 & 81.0 & 26.5 & 73.5 & 8.5 & 52.4 \\
\rowcolor{gray!20}
& CDAN \citep{Zhang2018CollaborativeAA} & 85.2 & 66.9 & 83.0 & \underline{50.8} & 84.2 & 74.9 & 88.1 & 74.5 & 83.4 & 76.0 & 81.9 & \underline{38.0} & 73.9 \\
\rowcolor{gray!20}
& DWL \citep{xiao2021dynamic} & 90.7 & 80.2 & 86.1 & 67.6 & 92.4 & 81.5 & 86.8 & 78.0 & 90.6 & 57.1 & 85.6 & \underline{28.7} & 77.1 \\
\rowcolor{gray!20}
& MEDM \citep{9537640} & 93.5 & 80.4 &\textbf{90.8} & 70.3 & 92.8 & 87.9 & 91.1 & 79.8 & 93.7 & 83.6 & 86.1 & 38.7 & 82.4 \\
\rowcolor{gray!20}
& SIDA \citep{DBLP:journals/corr/abs-2110-12184} & 95.4 & 83.1 & 77.1 & 64.6 & 94.5 &97.2 & 88.7 & 78.4 & 93.8 & 89.9 & 85.2 & 59.4 & 84.0 \\
\rowcolor{gray!20}
& CAN \citep{kang2019contrastive} & 97.0 & 87.2 & 82.5 & 74.3 & \textbf{97.8} & 96.2 & 90.8 & 80.7 & \textbf{96.6 }& \textbf{96.3} & 87.5 & 59.9 & 87.2 \\
\rowcolor{gray!20}
& CDCL \citep{wang2022cross} & \textbf{97.4} & \textbf{89.5} & 85.9 & \textbf{78.2} & 96.4 & 96.8 & \textbf{91.4} & 83.7 & 96.3 & 96.2& \textbf{89.7} & 61.6 & \textbf{88.6} \\
\rowcolor{gray!20}
\textbf{V. 17}& SDAT \citep{rangwani2022closer} & 95.8& 85.5 & 76.9 & 69.0 & 93.5 & \textbf{97.4} & 88.5 & 78.2 & 93.1 & 91.6 & 86.3 & 55.3 & 84.3\\
\rowcolor{gray!20}
& MIC\cite{Hoyer_2023_CVPR} & 96.7 & 88.5 & 84.2 & 74.3 & 96.0 & 96.3 & 90.2 & 81.2 & 94.3 & 95.4 & 88.9 & 56.6 & 86.9 \\
\rowcolor{gray!20}
& FPS & 94.3 & 86.2 & 81.3 & 71.9 & 95.7 & 96.9 & 90.4 & \textbf{84.3} & 92.8 & 92.5 & 83.5 & \textbf{63.4} & 86.1 \\

\hline
& source \citep{dosovitskiy2020image} & 97.7 & 48.1 & 86.6 & 61.6 & 78.1 & 63.4 & 94.7 & 10.3 & 87.7 & 47.7 & 94.4 & 35.5 & 67.1 \\
& CDTrans* \citep{xu2021cdtrans} & 97.1 & 90.5 & 82.4 & 77.5 & 96.6 & 96.1 & 93.6 & 88.6 & 97.9 & 86.9 & 90.3 & 62.8 & 88.4 \\
& TVT \citep{yang2023tvt} & 92.9 & 85.6 & 77.5 & \underline{60.5} & 93.6 & 98.2 & 89.4 & 76.4 & 93.6 & 92.0 & 91.7 & 55.7 & 83.9 \\
& SSRT \citep{sun2022safe} & 98.9 & 87.6 & \textbf{89.1} & 84.8 & 98.3 & 98.7 & 96.3 & 81.1 & 94.9 & 97.9 & 94.5 & \underline{43.1} & 88.8 \\
& SDAT \citep{rangwani2022closer} & 97.1& 88.4 & 80.9 & 75.3 & 95.4 & 97.9 & 94.3 & 85.5 & 95.8 & 91.0 & 93.0 & 65.4 & 88.4\\
& MIC\cite{Hoyer_2023_CVPR} & 99.0 & \textbf{93.3} & 86.5 & \textbf{87.6} & \textbf{98.9} & \textbf{99.0} & \textbf{97.2} & \textbf{89.8} & \textbf{98.9} & \textbf{98.9} & \textbf{96.5} & 68.0 & \textbf{92.8} \\
& FPS & \textbf{99.1} & 91.0 & 84.4 & 77.6 & 97.5 & 96.8 & 95.5 & 79.6 & 94.4 & 94.4 & 95.9 & \textbf{68.2}  & 89.5 \\

\midrule
\rowcolor{gray!20}
& source \citep{he2016deep} & 68.9 & 60.7 & 68.4 & 62.5 & - & - & - & - & - & - & - & - & 65.1 \\
\rowcolor{gray!20}
& CDAN \citep{Zhang2018CollaborativeAA} & 92.9 & 71.0 & 94.1 & 69.3 & - & - & - & - & - & - & - & - & 81.8 \\
\rowcolor{gray!20}
& SHOT \citep{Liang2020DoWR} & 94.0 & 74.3 & 90.1 & 74.7 & - & - & - & - & - & - & - & - & 83.3 \\
\rowcolor{gray!20}
 & MEDM\citep{9537640} &
95.5 & \textbf{78.5} & 92.8 & 77.2 & - & - & - & - & - & - & - & - & 86.0 \\
\rowcolor{gray!20}
& SIDA\citep{DBLP:journals/corr/abs-2110-12184} &
94.5 & 76.2 & 95.7 & 76.6 & - & - & - & - & - & - & - & - & 85.8 \\
\rowcolor{gray!20}
& CAN \citep{kang2019contrastive} & 94.5 & 77.0 & 95.0 & 78.0 & - & - & - & - & - & - & - & - & 86.1 \\
\rowcolor{gray!20}
\textbf{O. 31}& CDCL \citep{wang2022cross} & 96.0 & 75.5 & \textbf{96.0} & 77.2 & - & - & - & - & - & - & - & - & 86.2 \\
\rowcolor{gray!20}

& FPS & \textbf{96.6} & 77.4 & 95.1 & \textbf{78.4} & - & - & - & - & - & - & - & - & \textbf{86.4} \\
\hline
& source \citep{dosovitskiy2020image} & 90.4 & 76.4 & 90.8 & 76.8 & - & - & - & - & - & - & - & - & 83.6 \\
& CDTrans* \citep{xu2021cdtrans} & 97.0 & 81.9 & 96.7 & 81.1 & - & - & - & - & - & - & - & - & 89.2 \\
& TVT \citep{yang2023tvt} & 96.4 & \textbf{86.0} & 96.4 & 84.9 & - & - & - & - & - & - & - & - & 90.9 \\
& SSRT \citep{sun2022safe} & 97.7 & 82.2 & 98.6 & 83.5 & - & - & - & - & - & - & - & - & 90.5 \\

& FPS & \textbf{98.3} & 85.3 & \textbf{98.8} & \textbf{85.1} & - & - & - & - & - & - & - & - & \textbf{91.9} \\
\bottomrule
\end{tabular}
\end{table*}

\begin{figure*}
\centering
\includegraphics[width=7.0in, trim=0cm 0.8cm 0.1cm 0.1cm, clip]{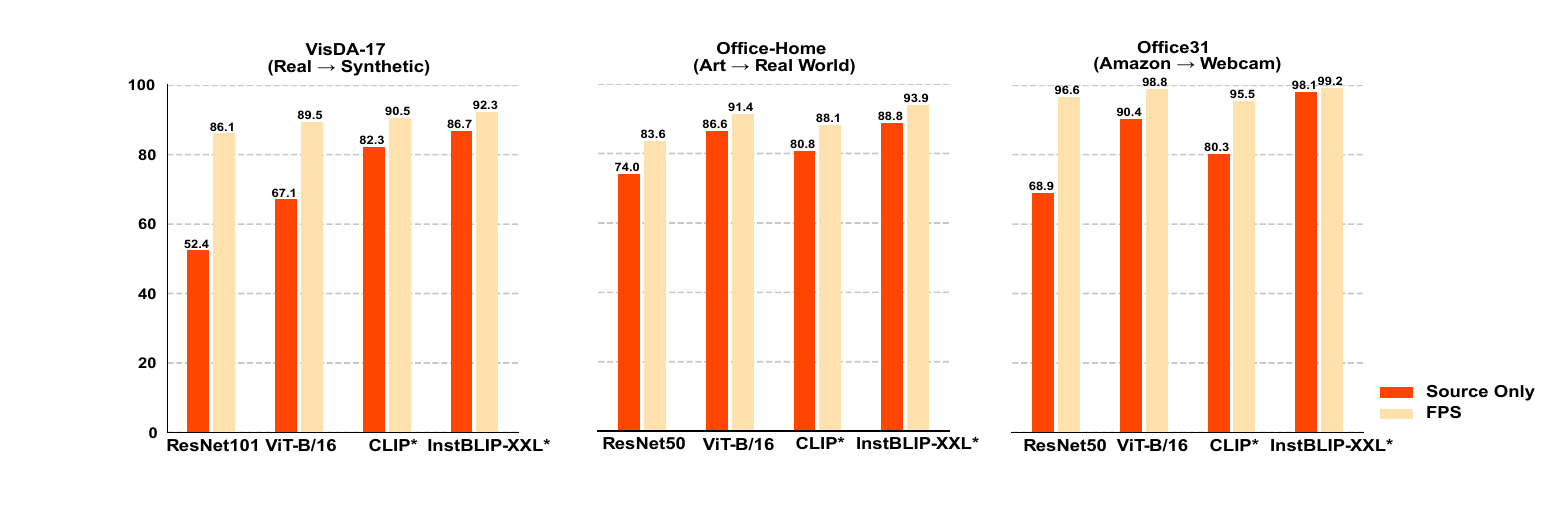}
\caption{ Across the three benchmarks, pure supervised accuracy and domain adaptation accuracy are compared for different pre-trained backbones. The source domain accuracy rates of CLIP* and InstructBLIP* are obtained from our experiments. The source domain accuracy rates of ResNet and ViT are from \cite{he2016deep} and \cite{dosovitskiy2020image}}
\captionsetup{font={scriptsize}} 
\label{R and performance b}
\end{figure*}

\subsubsection{Scalability to Foundation Models}
We further evaluated the performance of FPS using multimodal pretrained models—CLIP~\cite{Radford2021LearningTV} and InstructBLIP-XXL ~\cite{Dai2023InstructBLIPTG} (vision encoders), as backbone networks. CLIP, a pre-trained model not fine-tuned for classification tasks, demonstrates performance that is complementary yet competitive with the ViT model (Fig.~\ref{R and performance b}). InstructBLIP-XXL, a multimodal large language model (MLLM), incorporates a vision encoder that outperforms evaluated backbones  (Fig.~\ref{R and performance b}). Under pure source-domain supervision, this model achieves superior accuracy for the target domain, surpassing existing benchmarks and highlighting its enhanced feature extraction capabilities (Fig.~\ref{R and performance b}). This performance improvement arises in part from the model's larger parameter scale, which demands greater computational resources for conventional fine-tuning methods. Our approach requires only feature pre-extraction and adjustment of the final linear layer, meaning feature extraction is performed just once. Moreover, because no encoder gradients are required, inference remains stable even with very small batch sizes, which substantially reduces computational overhead.This efficiency enables the seamless integration of larger, state-of-the-art models with minimal overhead.

\subsection{Impact of Individual Components}

\begin{table*}[htbp]
\caption{\textbf{Representative ablations with vs. without feature-geometric degradation.} 
"Original" settings correspond to $(\alpha, \beta)=(1.0, 1.0)$, while "Degraded" refers to features distorted with $(\alpha, \beta)=(1.5, 1.5)$. 
For VisDA17, we report Macro-F1 and Mean per-class accuracy (C-Acc). 
The bold values indicate the best performance in each column under the respective degradation settings.}
\centering
\scriptsize
\setlength{\tabcolsep}{4.5pt}
\renewcommand{\arraystretch}{1.2}
\begin{tabular}{l|cc|cc|cc|cc}
\toprule
& \multicolumn{4}{c|}{Office-Home (Clipart$\rightarrow$Product, ViT-B/16)} 
& \multicolumn{4}{c}{VisDA17 (Syn$\rightarrow$Real, ResNet101)} \\
\cmidrule(lr){2-5}\cmidrule(lr){6-9}
Method 
& \multicolumn{2}{c|}{Original ($\alpha,\beta=1.0$)} 
& \multicolumn{2}{c|}{Degraded ($\alpha,\beta=1.5$)} 
& \multicolumn{2}{c|}{Original ($\alpha,\beta=1.0$)} 
& \multicolumn{2}{c}{Degraded ($\alpha,\beta=1.5$)} \\
& Acc & Macro-F1 & Acc & Macro-F1 
& Macro-F1 & C-Acc & Macro-F1 & C-Acc \\
\midrule
Baseline (SCE)
& 86.2 & 85.1 & 70.3 & 67.1
& 65.5 & 67.7 & 27.0 & 28.7 \\
$+\mathcal{L}_{\mathrm{CR}}$ only
& 86.3 & 85.3 & 70.4 & 67.3
& 65.7 & 67.7 & 26.4 & 28.2 \\
$+\mathcal{L}_{\mathrm{SE}}$ only
& 78.3 & 72.7 & 64.9 & 59.2
& 62.0 & 64.4 & 22.5 & 25.0 \\
$+\mathcal{L}_{\mathrm{CE}}$ only
& 87.6 & 86.7 & 70.6 & 68.8
& 68.2 & 70.2 & 29.4 & 30.4 \\
\midrule
Entropy core ($\mathcal{L}_{\mathrm{SE}}+\mathcal{L}_{\mathrm{CE}}$)
& 90.1 & 88.4 & 77.9 & 74.3
& 82.8 & 84.2
& 51.1 & 51.1 \\
Core $+\mathcal{L}_{\mathrm{CR}}$
& 90.2 & 88.5 & 78.0 & 74.4
& 84.3 & 85.2
& 53.9 & 53.3 \\
Core $+\mathcal{L}_{\mathrm{CR}}$+margin (All)
& \textbf{91.8} & \textbf{91.3} & \textbf{80.6} & \textbf{78.9}
& \textbf{84.7} & \textbf{86.1}
& \textbf{59.3} & \textbf{60.3} \\
\bottomrule
\end{tabular}
\vspace{1.5mm}

\label{tab:geom_ablation_main_colored}
\end{table*}

Table~\ref{tab:geom_ablation_main_colored} details the contribution of each component under both original and degraded feature geometries. On the original features, the entropy core ($\mathcal{L}_{\mathrm{SE}}+\mathcal{L}_{\mathrm{CE}}$) serves as a primary contributor to the observed performance gains. Specifically, Macro-F1 improves from 85.1\% to 88.4\% on Clipart$\rightarrow$Product, while VisDA-2017 exhibits substantial gains, with Macro-F1 rising from 65.5\% to 82.8\% and C-Acc from 67.7\% to 84.2\%. The consistent improvement in both metrics suggests that the method promotes more balanced performance across categories rather than biasing toward majority classes.

The importance of the dual-entropy mechanism is further reflected in the relatively smaller improvements observed when using $\mathcal{L}_{\mathrm{SE}}$, $\mathcal{L}_{\mathrm{CE}}$, or $\mathcal{L}_{\mathrm{CR}}$ in isolation compared to the baseline. This suggests that the performance gains are associated with the complementary interaction between sample-wise and category-wise regularization.

The entropy core remains consistently robust even under the severe geometric degradation described in Section~\ref{subsec:geometry_stress}. In this ablation, adding the consistency regularizer $\mathcal{L}_{\mathrm{CR}}$ and the margin-aware confidence weighting further improves performance, indicating that both components help stabilize the decision boundary under challenging geometric distortions, especially when the cross-domain shift is large and class separability is low. Among them, the margin-aware confidence weighting yields the most noticeable gain, as shown in Table~\ref{tab:geom_ablation_main_colored}. We further compare the proposed one with vanilla margin-aware weighting methods in Section \ref{app:MACW} of the Appendix. The results show that our method is more robust and brings more noticeable improvements on part of tasks.

Comprehensive results across the full $(\alpha,\beta)$ spectrum for original features are provided in Tables~\ref{tab:Append_grid_clipart2product} and~\ref{tab:Append_grid_visda} in the Appendix.

\subsection{Efficacy of Fixed-Encoder Adaptation}\label{fint_abl}
\begin{table*}[t]
\centering
\caption{Controlled comparison of different adaptation extents on Art$\rightarrow$Product (ResNet-50) under the same optimizer family, schedule family, and training budget. To ensure fairness in comparison with all batch-wise methods, FPS$^{*}$ represents the FPS variant that does not use global statistical metrics such as Margin-Aware Confidence Weighting and Density Weighting Strategy.}
\label{tab:extent_controlled_comparison}
\setlength{\tabcolsep}{6pt}
\renewcommand{\arraystretch}{1.15}
\begin{tabular}{lcccc}
\toprule
\textbf{Mode} & \textbf{Trainable Params}  & \textbf{Data Input Mode} & \textbf{Source Acc. (\%)} & \textbf{Target Acc. (\%)}\\
\midrule
Frozen (head-only) & 266,370  & Batch-wise & 99.13& 79.97 \\
Last-block         & 15,231,106 & Batch-wise & \textbf{99.51}  & 79.32\\
Full fine-tuning   & 25,823,402 & Batch-wise & 99.42 & 77.37 \\
\midrule
FPS$^{*}$          & 266,370  & Single-step & 99.15& \textbf{80.15} \\
\bottomrule
\end{tabular}
\end{table*}

To examine how the extent of adaptation affects performance under a fixed pretrained backbone (ResNet-50) and matched optimization settings, we conduct a controlled comparison among Frozen (Only the classifier can be trained), Last-block fine-tuning, Full fine-tuning, and FPS. The results on the Office-Home Art$\rightarrow$Product task and Product$\rightarrow$Art are reported in Table~\ref{tab:extent_controlled_comparison} and Table~\ref{tab:controlled_complexity_pa} in Appendix~\ref{app: ablation FT}. For each mode, the hyperparameters were determined via hyperparameter search; the detailed settings and sensitivity analyses are provided in Table~\ref{tab:hyperparam_settings} and Fig.~\ref{fig:fint_hyp} of the Appendix.

As the number of trainable parameters increases, source-domain accuracy improves slightly, indicating stronger fitting to the source data. However, this increase in model capacity does not translate into improved target-domain performance. In fact, both Last-block and Full fine-tuning exhibit lower target accuracy compared to the Frozen setting.

This result suggests that, under strong pretrained representations, simply enlarging the hypothesis space does not reliably improve cross-domain generalization under the same optimization protocol (see Appendix~\ref{app: ablation FT} for detailed settings). 
This trend is consistent with the source-fit/complexity trade-off implied by Theorem~\ref{thm:target_risk}: once empirical source risk is already low, expanding the hypothesis space without sufficient constraint can increase generalization difficulty more readily than it improves target-domain performance. 
In addition, FPS differs from standard batch-wise fine-tuning in that it performs single-step optimization, which may contribute to the empirical gains.

We note that this observation does not preclude the effectiveness of feature fine-tuning in general. In particular, additional regularization or constraints may mitigate the effect of increasing parameter scale. Regularization terms that improve feature robustness, especially those that cannot be optimized solely through classifier-only adaptation, such as the contextual consistency used in MIC~\cite{Hoyer_2023_CVPR},which can provide additional benefits in some tasks.

\subsection{Robustness Analysis}
\label{sec:robustness}

We evaluate the robustness of FPS under a series of controlled stress conditions
that systematically perturb the target domain while keeping the feature encoder fixed.
The objective is to assess whether the gains of FPS persist when key assumptions
about feature geometry, label distribution, and label-space consistency are violated.

\subsubsection{Robustness to Feature Geometry Degradation}
\label{subsec:geometry_stress}

We first examine robustness under controlled degradation of the target feature geometry induced by a fixed, pre-trained encoder. Since the classifier is trained on frozen representations, the attainable target performance is fundamentally constrained by the class-conditional structure of the feature space and the degree of cross-domain alignment.

To explicitly control these factors, we construct synthetically degraded target features along two interpretable axes: intra-class dispersion and cross-domain class-wise shift. Let $\mu_s^c$ and $\mu_t^c$ denote the source and target feature centers for class $c$. For each target feature $x^i_t$ with label $y_t^i = c$, we define
\begin{equation}
x^i_t(A,B)
= \mu_s^c + \underbrace{B(\mu_t^c - \mu_s^c)}_{\text{cross-domain shift}}
+ \underbrace{A(x^i_t - \mu_t^c)}_{\text{intra-class dispersion}},
\label{eq:affine_degradation}
\end{equation}
where $A (\ge 1)$ increases intra-class dispersion and $B (\ge 1)$ amplifies cross-domain class-wise misalignment. Specifically, $A$ scales the intra-class distances within the target domain by a factor of $A$, while $B$ scales the distances between class centers across the source and target domains by a factor of $B$. These parameters respectively influence the assumption of intra-class compactness and the extent of domain shift. The original features correspond to $(A,B)=(1,1)$. Notably, increasing $A$ (larger intra-class dispersion) tends to raise pseudo-label noise by causing stronger class overlap and less confident predictions.

\begin{figure}
\centering
\includegraphics[width=3.3in]{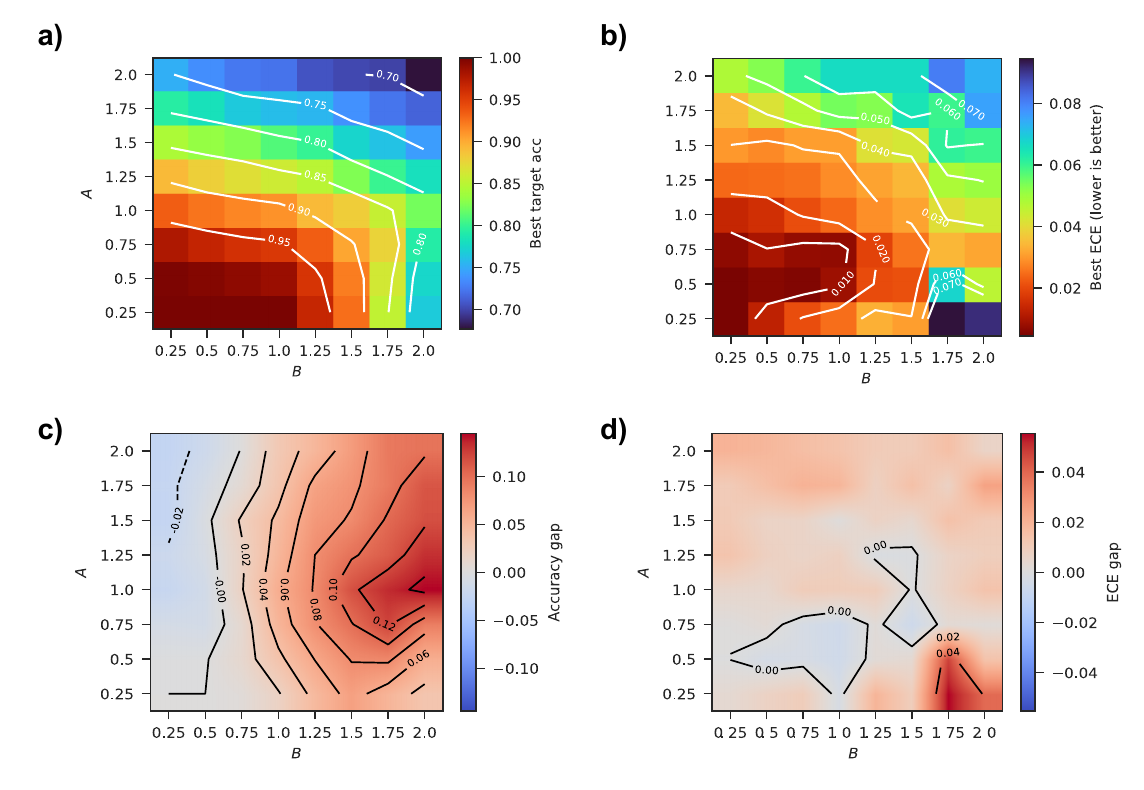}
\caption{\textbf{Class-conditional affine transformation to control intra-class dispersion and cross-domain shift, parameterized by $(A,B)$ (Eq.~\ref{eq:affine_degradation}), on Art$\rightarrow$Product.}
(a) Best target accuracy and (b) best ECE (lower is better) of FPS over the $(A,B)$ grid.
ECE follows the standard definition of~\cite{guo2017calibration} (Appendix \ref{app:ece}).(c) and (d) represent the gap maps comparing accuracy and ECE to Source-Only, respectively.}
\captionsetup{font={scriptsize}}
\label{fig:stress_art2prod}
\end{figure}

Figure~\ref{fig:stress_art2prod} illustrates the performance of FPS on the Art$\rightarrow$Product task under different degradation levels. 
As the parameters $(A,B)$ increase, both transfer accuracy and calibration quality decline, reflecting progressive deviations from the representation prior established by the frozen encoder. 
Nevertheless, FPS consistently outperforms the Source-Only baseline. Notably, different regions of the $(A,B)$ grid reveal distinct behaviors: when $B$ increases under small-to-moderate $A$ (dominant cross-domain shift), FPS still yields strong optima, indicating effective boundary re-alignment in a largely preserved geometry; when $A$ becomes large (dominant intra-class dispersion), both accuracy and ECE degrade more sharply, consistent with separability being fundamentally weakened in the fixed space. 

The gap maps in Fig.~\ref{fig:stress_art2prod}c–d further show that FPS improves accuracy while keeping similar ECE over most of the $(A,B)$ grid, suggesting that its performance gains are not accompanied by systematic overconfidence, particularly under moderate-to-strong distortions. Two exceptions are also consistent with the geometric interpretation: 
(i) within the low-$B$ regime (characterized by a small cross-domain shift), the accuracy gap becomes negligible (and can even be marginally negative). This is because the Source-Only boundary is already nearing optimality, and the supplementary regularization offers limited scope for enhancement; 
(ii) at the corner where $B$ is extremely large yet $A$ is small (representing tight clusters with a strong class-wise shift), the ECE gap can become slightly positive. This indicates that, in this extreme scenario, the solution with the highest accuracy might lead to sharpened predictions and a mild degradation in calibration. The comprehensive stress-test results for VisDA-2017 and OfficeHome, encompassing full $(A,B)$ grid evaluations, are presented in Appendix~\ref{app:add res for geo deg}. These results display the same qualitative tendency, further validating the robustness of FPS for decision boundary optimization.

\emph{Robustness of the ICDM-Based Selector.}
Beyond classification performance, we examine whether the ICDM metric
remains a reliable unsupervised hyperparameter selector under feature
geometry degradation. The details are provided in Appendix~\ref{app:icdm}.
Our analysis reveals a clear asymmetry:
the selector is robust to cross-domain class-wise shifts ($B$),
but degrades as intra-class dispersion ($A$) increases.
This behavior aligns with the metric’s reliance on within-class compactness.

Intuitively, enlarging $A$ weakens within-class compactness and increases class overlap, which directly amplifies pseudo-label noise and consequently degrades the estimated intra-class distances; we acknowledge that ICDM can deteriorate under sufficiently high noise.
Nevertheless, the selector remains robust in a broad range of moderate degradations and is notably insensitive to class-wise shifts($B$), since $B$ mainly shifts class centers across domains while preserving within-class structure, keeping pseudo-label assignments comparatively stable. 

\subsubsection{Robustness under Target-Domain Class Imbalance}
\label{subsec:imbalance_stress}

We next evaluate robustness under target-domain class imbalance,
a pervasive challenge in real-world deployment that biases optimization
toward high-frequency classes and exacerbates class-wise risk disparity.

Using the VisDA-2017 dataset, we construct multiple imbalance regimes ranging from
balanced to heavily long-tailed distributions.
Rather than introducing imbalance synthetically from scratch,
we reweight the intrinsic class frequencies to obtain controlled and interpretable
stress patterns.
All imbalance constructions are summarized in Appendix~\ref{app:imbalance_setup}.

Figure~\ref{fig:class_imbalanced_summary} in Appendix~\ref{app:imbalance_setup} shows that while the gains in Top-1 accuracy
remain relatively modest, FPS consistently yields substantially larger improvements
in Macro-F1 across all imbalance regimes.
This discrepancy indicates that class imbalance primarily amplifies
class-wise error concentration, which is effectively mitigated by FPS. 

As shown in Fig. \ref{fig:difference_w_factor} in Appendix, 
we observed that the introduction of class-balanced weighting (introduced in  \hyperref[class_weight]{\textit{\nameref{class_weight}}}) contributes to improved performance in scenarios characterized by class imbalance. 
This indicates that it is an effective method for mitigating the impact of class imbalance.

Overall, these results demonstrate that FPS shifts the learning focus from a
frequency-dominated empirical risk toward a more class-equitable objective,
leading to improved robustness for low-frequency and disadvantaged classes.
Per-class analyses and additional quantitative results are provided in
Appendix~\ref{app:imbalance_setup}.

\subsubsection{Robustness under Label-Space Mismatch}
\label{subsec:label_mismatch}
In practical settings, the assumption of a perfectly shared label space
between source and target domains is often violated.
To assess robustness under such conditions,
we simulate partial label-space mismatch by removing subsets of target classes
while keeping the classifier output dimension fixed.

\begin{figure}
\centering
\includegraphics[width=3.5in,trim=0cm 0.15cm 0cm 0.25cm, clip]{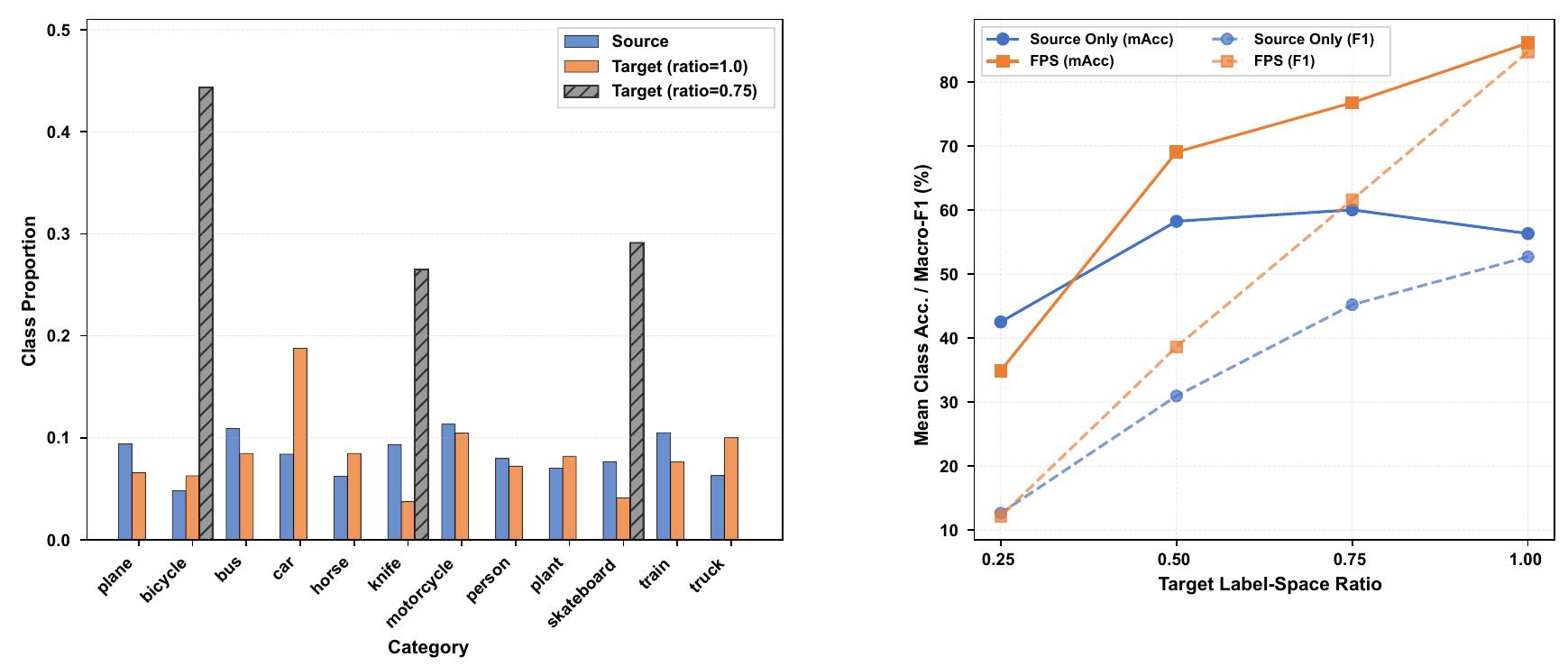}
\caption{\textbf{Robustness of FPS under head label-space mismatch.}
Left: label frequency distributions for the source, full target, and head-mismatched target domains.
  Right: evolution of Top-1 accuracy and Macro-F1 as the retained target label-space proportion diminishes.}
\label{fig:label_mismatch_head}
\end{figure}

Across multiple removal patterns, FPS exhibits significantly more graceful
performance degradation than the baseline (Only source),
particularly in terms of Macro-F1.
As shown in Fig.~\ref{fig:label_mismatch_head}, the head-mismatch setting induces a pronounced shift in the target label-frequency profile (left),
and the right panel traces how both Top-1 accuracy and Macro-F1 evolve as the retained target label-space ratio decreases.
While both methods deteriorate when the label space becomes severely incomplete, we acknowledge that in the extreme regime (e.g., ratio $=0.25$)
the adaptation signal becomes too sparse and the performance can collapse (and may even lose the advantage over the baseline).
Nevertheless, for the practically relevant range of moderate mismatch (e.g., ratios $\ge 0.5$),
FPS consistently maintains a higher Macro-F1 and degrades more smoothly than Source-Only, indicating improved class-equitable robustness.
This behavior suggests that FPS does not critically rely on label-space completeness;
instead, by enforcing structural regularization on target predictions, it stabilizes decision-plane placement under missing-class scenarios,
until the representation provides too little class structure to constrain the boundary.
Results under additional mismatch patterns (random and tail) are reported in
Appendix~\ref{app:label_mismatch}.

\subsection{Applications in Diverse Real-World Scenarios}

\begin{figure}
\centering
\includegraphics[width=3.3in]{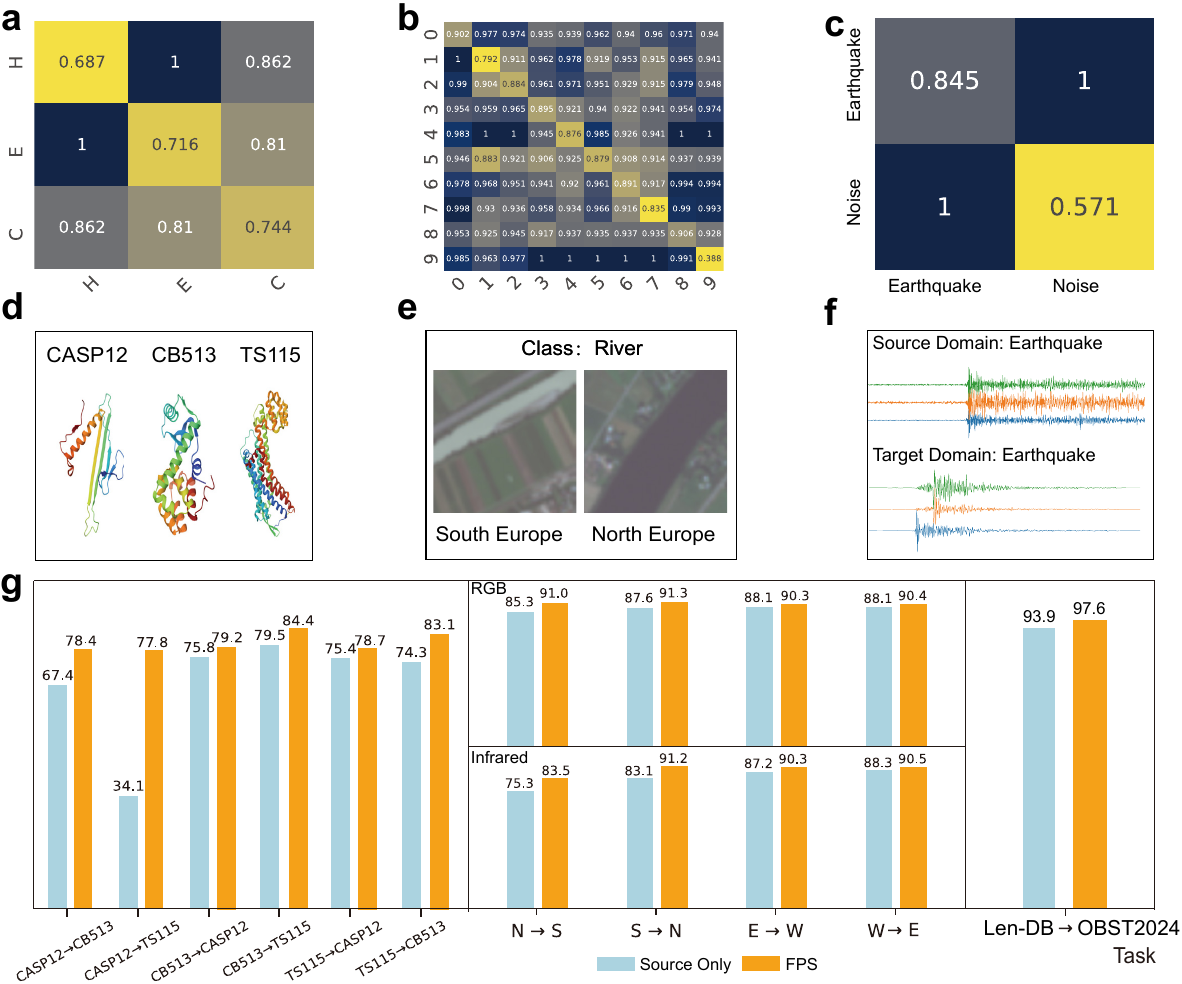}
\caption{Results for real-world applications. (a/b/c) illustrate the average distance between samples in the target domains (analogous to Figure \ref{feature distribution pattern b}) for protein structure prediction, remote sensing image classification, and earthquake event detection, respectively. (d/e/f) provide representative examples corresponding to (a/b/c), respectively. (g) presents the domain-adaptation results, with the horizontal axis representing distinct domain adaptation tasks. The three distinct sections, demarcated by solid vertical lines, correspond to the three applications represented in (a/b/c). In the second section of (g), N, S, E, and W denote North, South, East, and West, respectively, corresponding to four distinct geographical regions in Europe.}
\captionsetup{font={scriptsize}} 
\label{application}
\end{figure}

Since FPS operates on pre-extracted features, it is naturally applicable beyond conventional vision benchmarks. To demonstrate this versatility, we evaluated FPS in several classic scenarios using diverse empirical datasets \cite{helber, magriniLocalEarthquakesDetection2020, niksejelOBSTransformerDeeplearningSeismic2024, Klausen2019}. The overall results and representative examples are shown in Fig.~\ref{application}. Detailed information is provided in \ref{app: ESAT_RGB}, \ref{app: Eq Det}, \ref{app: Prot Struct}  and  \ref{app: real-world hypersetting} in the Appendix.

Protein secondary structure prediction is a critical challenge in computational biology and bioinformatics. This task involves mapping each input amino acid to a label in the set \(\{ \text{Helix (H)}, \text{Strand (E)}, \text{Other (C)} \}\), representing a sequence-to-sequence prediction problem. Despite significant advances in prediction methods, the diversity and complexity of protein structures remain formidable challenges. Neural networks trained on limited datasets often struggle to generalize to novel protein structures. Using the open dataset \cite{Klausen2019}, the domain adaptation task involves transferring knowledge between pairs of datasets from CB513, CASP12, and TS115. For feature extraction, we employ the esm2\_t33\_650M\_UR50D\cite{Lin2022.07.20.500902} model. The extracted feature distribution exhibits clear intra-class compactness and inter-class separability (Fig.~\ref{application}a), leading to a significant improvement in the Q3 accuracy rate of the target domain (Fig.~\ref{application}g).

Remote sensing image recognition is a fundamental task in environmental monitoring and geospatial analysis, and numerous intelligent recognition methods have been developed \cite{9127795}. However, due to factors such as vegetation coverage and developmental disparities, neural networks trained on limited datasets often exhibit poor generalization when applied to new regions. Our application focuses on identifying specific land cover and land use features from remote sensing imagery. Using the EuroSAT\_MS dataset \cite{helber}, the domain adaptation task involves transferring knowledge from one region of Europe to another, such as from southern to northern regions. The feature encoder employs the same pre-trained ViT model used in the benchmark test. As illustrated in Fig. \ref{application}b, the feature distribution exhibits clear intra-class compactness and inter-class separability, enabling effective discrimination. After adaptation, average recognition accuracy improves from 87.23\% to 90.78\% in the RGB band and from 83.4\% to 88.9\% in the infrared band (Fig. \ref{application}g). Notably, the pre-trained ViT models, despite not being trained on infrared band imagery, can extract meaningful and effective classification features.

Earthquake event detection serves as a fundamental task in seismic activity monitoring, and numerous intelligent detection methods have already been developed \cite{perol2018convolutional,zhu2018phasenetdeepneuralnetworkbasedseismicarrival}. However, due to the influence of observational environments, instrumentation, and other factors, neural networks trained on limited datasets often struggle to ensure generalizability when applied to new data \cite{zhuSeismicArrivaltimePicking2023}. The applied domain adaptation task involves transferring knowledge from an global dataset, Len-DB \cite{magriniLocalEarthquakesDetection2020}, to an offshore OBS dataset OBST2024 ~\cite{niksejelOBSTransformerDeeplearningSeismic2024}. The feature encoder is PhaseNet~\cite{zhu2018phasenetdeepneuralnetworkbasedseismicarrival}, which is pretrained on the phase-picking task using two open seismic datasets \cite{mousavi2019earthquake,munchmeyer2022seisbench}. Using this encoder, the feature distribution exhibits intra-class clustering and inter-class separation patterns, which can be theoretically distinguished (Fig.~\ref{application}c) and the detection precision improves from 93.90\% to 97.57\% after adaptation (Fig.~\ref{application}g).

\section{DISCUSSION and CONCLUSION}

In this work, we introduce the Feature-space Planes Searcher (FPS), a domain adaptation framework that optimizes decision planes within frozen pre-trained feature spaces using structural priors. Our empirical observations suggest that cross-domain performance degradation can often be alleviated by correcting decision-boundary misalignment within a frozen feature space, without necessarily modifying the feature representations. To achieve this, we propose unsupervised optimization objectives—target conditional entropy minimization, category entropy maximization, and stochastic pooling-based consistency regularization—to enable effective boundary estimation. FPS provides an alternative perspective to conventional domain adaptation approaches that rely on feature alignment or confusion, reformulating domain adaptation as refining the final decision boundary within a frozen feature space.  This approach highlights the strong transferability of pre-trained model features and facilitates quantitative analysis of domain shift constraints.

\bibliographystyle{IEEEtran}
\bibliography{sn-bibliography}

\begin{IEEEbiography}[{\includegraphics[width=1in,height=1.25in,clip,keepaspectratio]{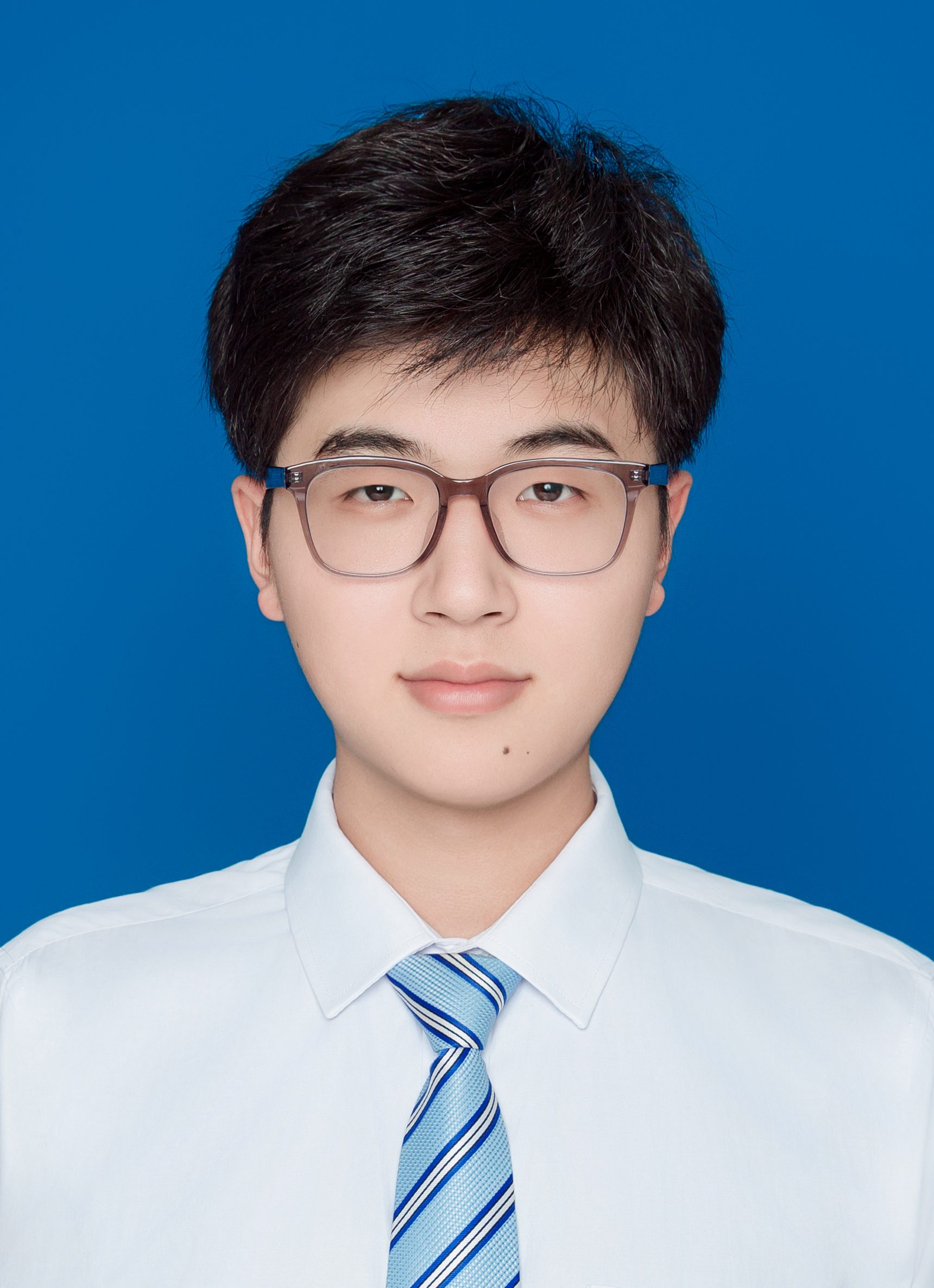}}]{Zhitong Cheng}
 received the B.Sc. degree in Computational Mathematics from Harbin Institute of Technology in 2024 and is currently pursuing the Ph.D. degree in Applied Mathematics at the same institution. 
\par His research interests include statistical machine learning , with a particular focus on transfer learning. He is also interested in diffusion modeling theory and its applications to seismic prestack data processing.  
\end{IEEEbiography}

\begin{IEEEbiography}[{\includegraphics[width=1in,height=1.25in,clip,keepaspectratio]{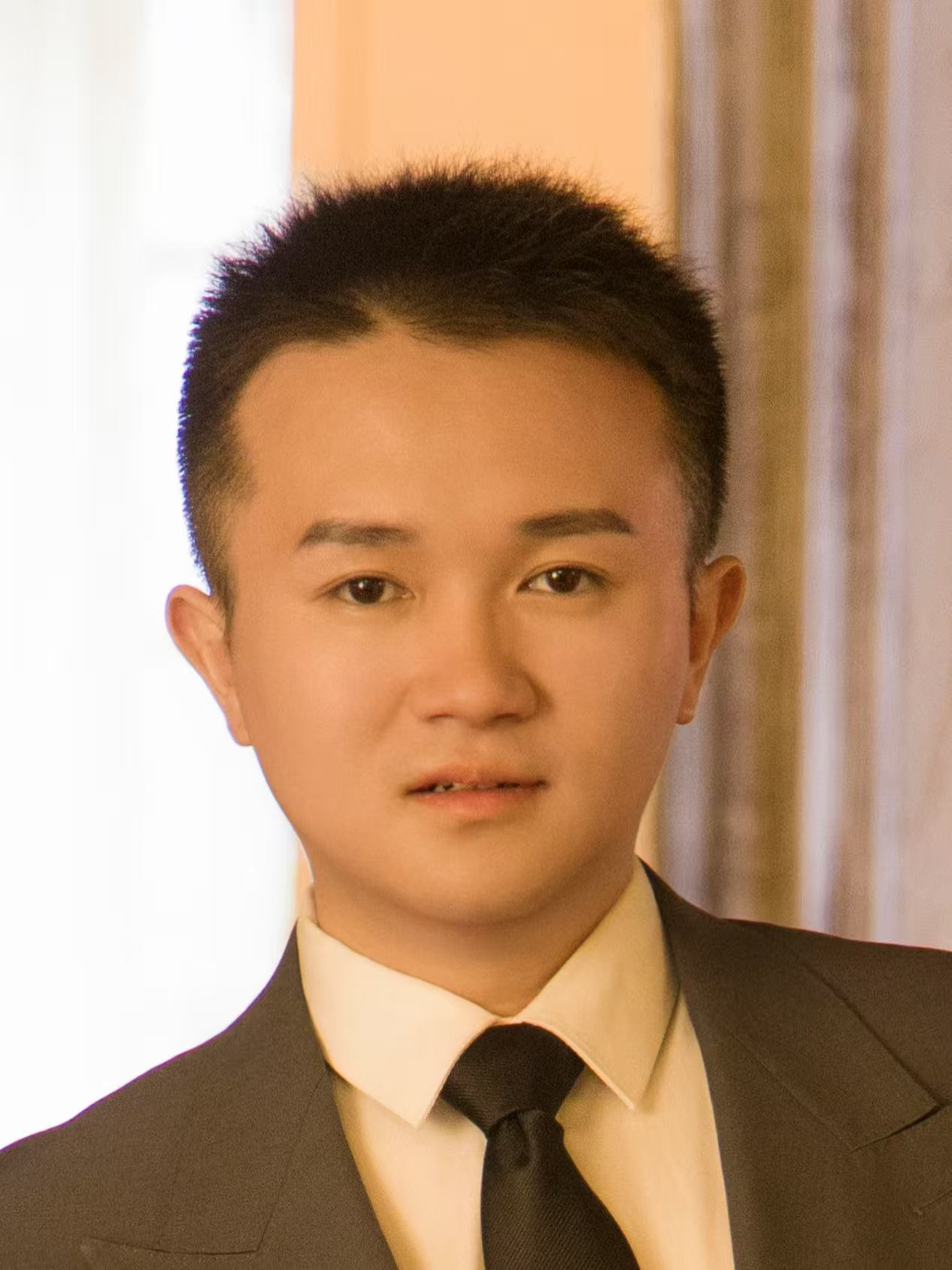}}]{Yiran Jiang}
received his B.Sc. in Geophysics in 2017 and his Ph.D. in Solid Earth Geophysics in 2023, both from the School of Earth and Space Sciences at Peking University. He is working as an Associate Researcher at the School of Mathematics, Harbin Institute of Technology. His research interests lie in geophysics, artificial intelligence, and their interdisciplinary applications.  
\end{IEEEbiography}

\begin{IEEEbiography}[{\includegraphics[width=1in,height=1.25in,clip,keepaspectratio]{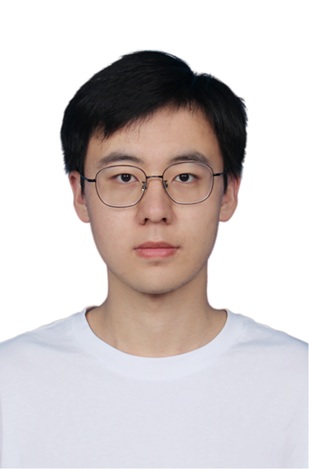}}]{Yulong Ge}
received the B.Sc. degree in computational mathematics from Harbin Institute of Technology in 2024, and is currently pursuing the MSc degree in applied mathematics at the same institution. His research interests lie in the theory and applications of diffusion generative models, neural network interpretability with both theoretical foundations and empirical explanations, as well as the mathematical analysis of deep learning systems.  
\end{IEEEbiography}

\begin{IEEEbiography}[{\includegraphics[width=1in,height=1.25in,clip,keepaspectratio]{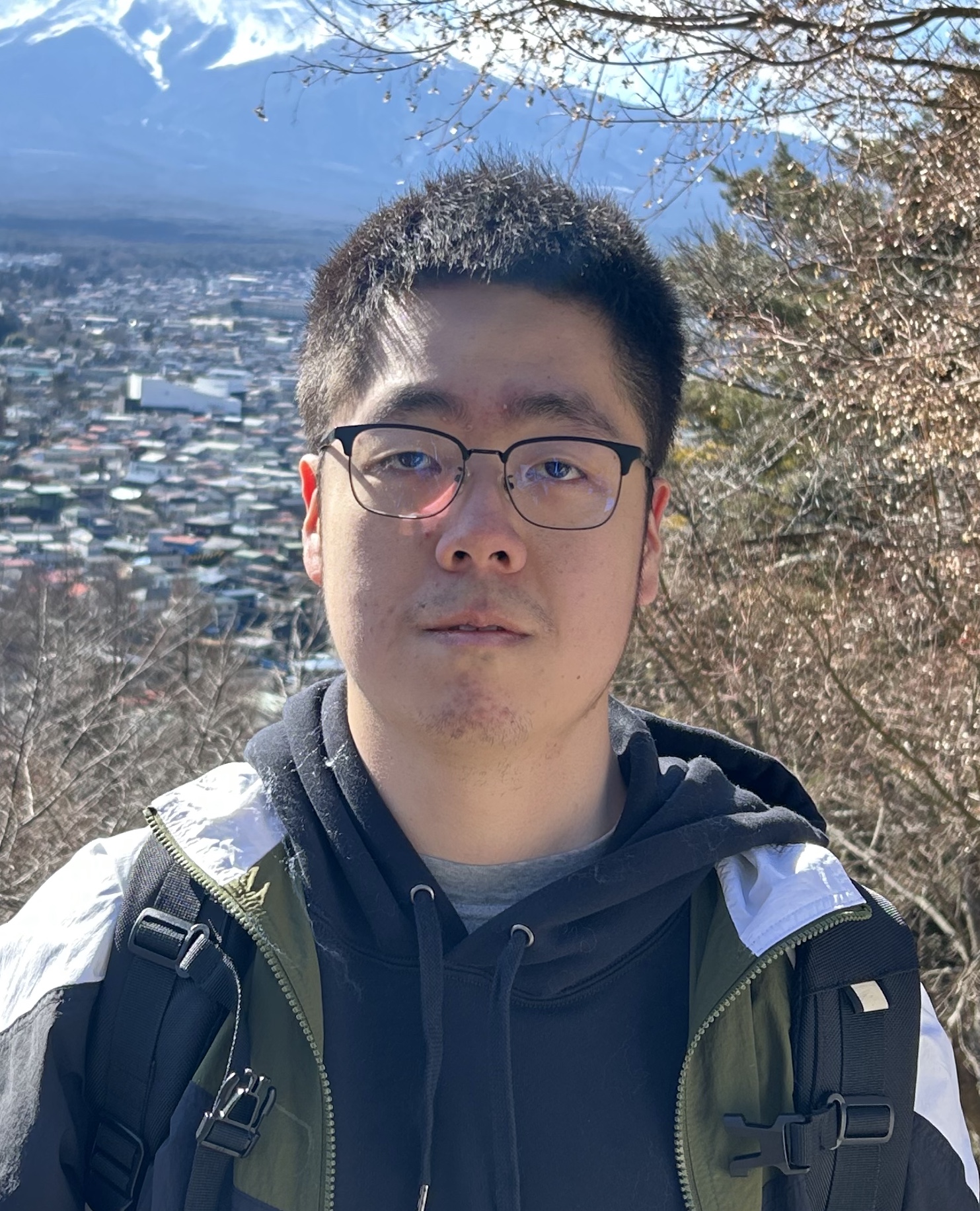}}]{Yufeng Li}
received the B.Sc. degree in Information and Computing Science from Northeastern University, China, in 2024. He is currently pursuing the M.Sc. degree in Applied Mathematics at Harbin Institute of Technology, China. His research interests include geophysics and computer vision, with a particular focus on their intersection in geophysical applications. 
\end{IEEEbiography}

\begin{IEEEbiography}[{\includegraphics[width=1in,height=1.25in,clip,keepaspectratio]{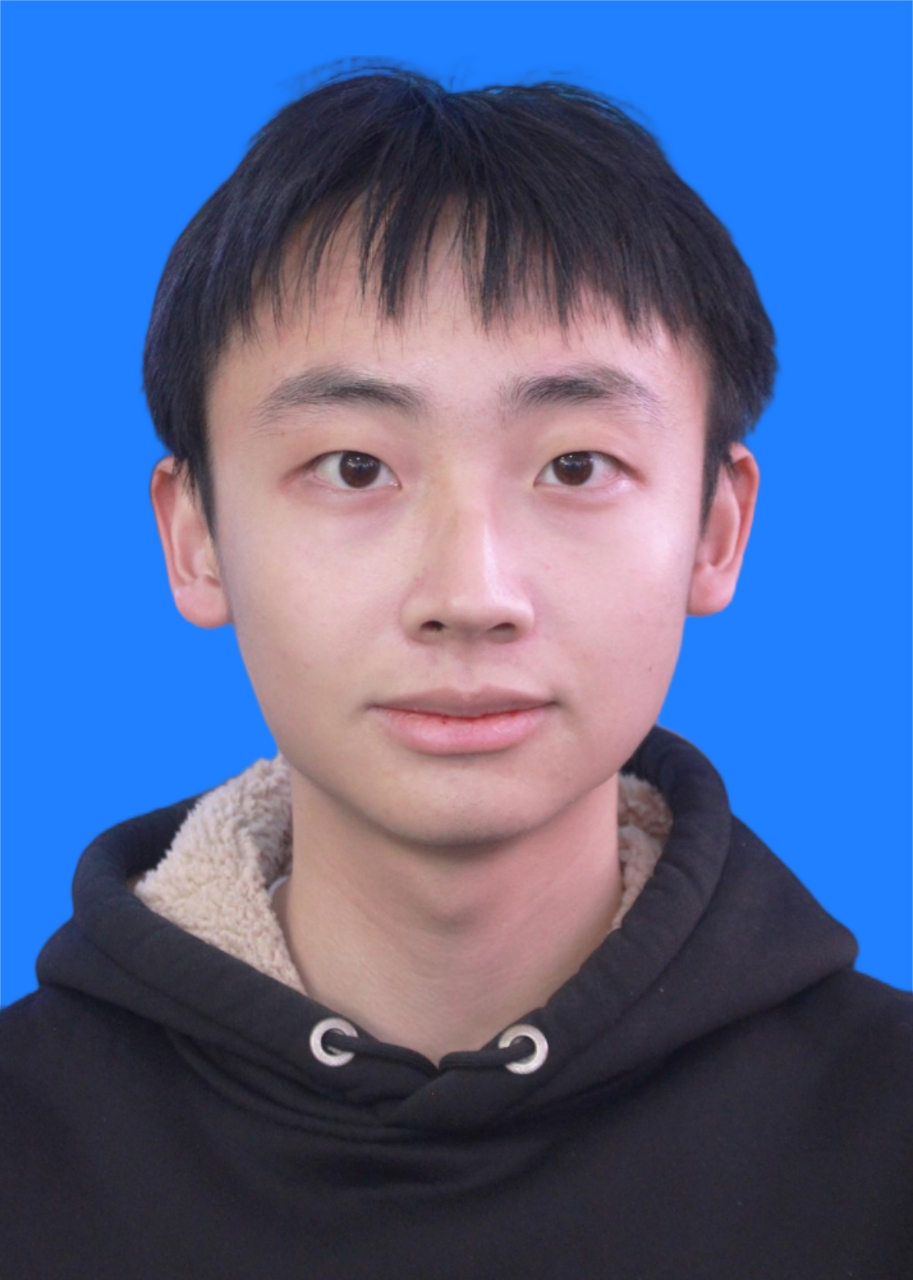}}]{Zhongheng Qin}
received the B.Sc. degree in mathematics and applied mathematics from Harbin Institute of Technology, in 2024.He is currently pursuing the master's degree in School of Mathematics, Harbin Institute of Technology. His research instersets include machine learning and deep learning, specifically the application in seismic images. 
\end{IEEEbiography}

\begin{IEEEbiography}[{\includegraphics[width=1in,height=1.25in,clip,keepaspectratio]{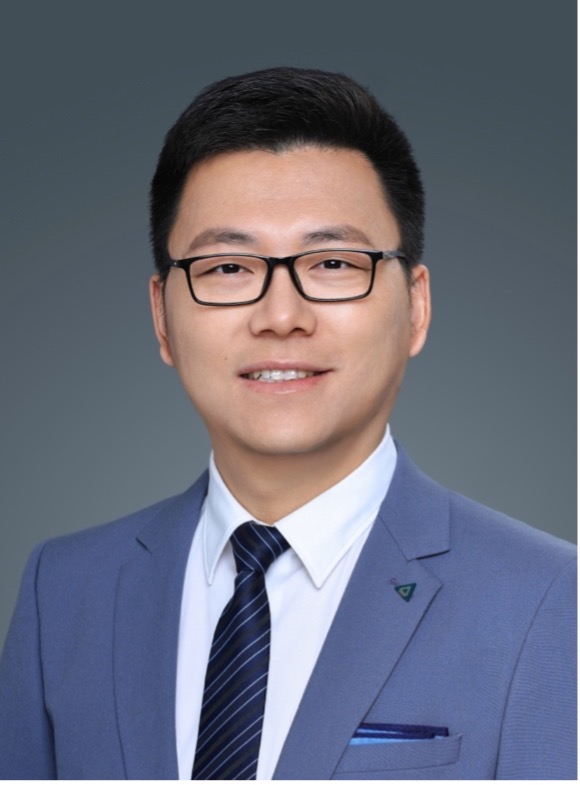}}]{Rongzhi Lin} received his B.Sc. in Geophysical Prospecting (2012) and M.Sc. (2015) from China University of Petroleum, Beijing, and a second M.Sc. in Geoscience from Khalifa University (formerly the Petroleum Institute), Abu Dhabi, UAE, in 2016. He earned his Ph.D. in Geophysics from the University of Alberta in 2022. He is currently an Associate Professor at Harbin Institute of Technology. His research interests include signal analysis, inverse problems, optimization, simultaneous source acquisition, and machine learning for geophysical exploration.
  
\end{IEEEbiography}

\begin{IEEEbiography}[{\includegraphics[width=1in,height=1.25in,clip,keepaspectratio]{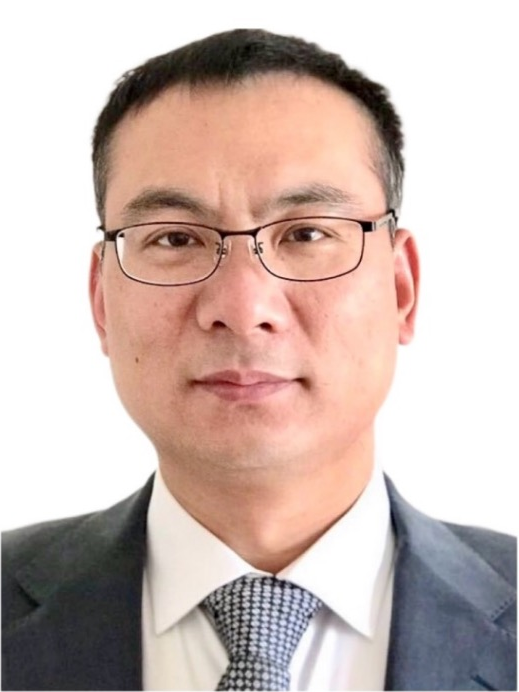}}]{Jianwei\_Ma}
(SeniorMember,IEEE) received the B.Sc. degree in engineering mechanics from Dalian University of Technology in 1998, and the PhD degree in solid mechanics from Tsinghua University in 2002. He is currently a Boya Distinguished Professor at the School of Earth and Space Sciences, Peking University, and the Director of the AI for Earth Science Center and the Smart Habitable Planet Center at the Institute for Artificial Intelligence, Peking University. His research interests include seismic signal processing, inverse problems, sparse representation, and deep learning methods for geophysical data. He has led several national key R\&D projects and published extensively in top-tier journals. He is a recipient of the National Science Fund for Distinguished Young Scholars and the Ten Thousand Talents Program, and serves as an Associate Editor for the IEEE Transactions on Geoscience and Remote Sensing.
\end{IEEEbiography}

\end{document}